\documentclass{article}
\usepackage{arxiv}

\usepackage[utf8]{inputenc} 
\usepackage[T1]{fontenc}    
\usepackage{hyperref}       
\usepackage{url}            
\usepackage{booktabs}       
\usepackage{amsfonts}       
\usepackage{nicefrac}       
\usepackage{microtype}      
\usepackage{lipsum}		
\usepackage{graphicx}
\usepackage[numbers]{natbib}
\usepackage{doi}

\usepackage{orcidlink}
\usepackage[title]{appendix}%

\newcommand{\fvx}{{FovEx}}
\usepackage{amsmath}
\DeclareMathOperator{\R}{\mathbb{R}}
\usepackage{newtxmath}

\usepackage{pgfplots}
\usepgfplotslibrary{groupplots}

\title{FovEx: Human-inspired Explanations for Vision Transformers and Convolutional Neural Networks}

\author{ \href{https://orcid.org/0000-0002-2447-9656}{\includegraphics[scale=0.06]{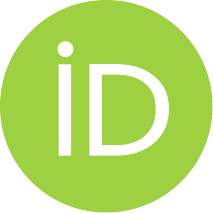}\hspace{1mm}Mahadev Prasad Panda} \\
	Department AIBE,\\
	FAU Erlangen-N\"urnberg,\\
	Erlangen, Germany \\
	\texttt{mahadev.prasad.panda@fau.de}
\And
\href{https://orcid.org/0000-0002-9133-8669}{\includegraphics[scale=0.06]{orcid.pdf}\hspace{1mm}Matteo Tiezzi} \\
	PAVIS,\\
	Istituto Italiano di Tecnologia (IIT),\\
	Genova, Italy \\
	\texttt{matteo.tiezzi@iit.it}
 \And
\href{https://orcid.org/0000-0002-1097-8534}{\includegraphics[scale=0.06]{orcid.pdf}\hspace{1mm}Martina 
 Vilas} \\
	Ernst Strüngmann Institute \\ for Neuroscience, \\
	Frankfurt, Germany \\
	\texttt{martinagonzalezvilas@gmail.com}
 \And
\href{https://orcid.org/0000-0002-6439-8076}{\includegraphics[scale=0.06]{orcid.pdf}\hspace{1mm}Gemma Roig} \\
	Goethe-Universität Frankfurt am Main \\
	Frankfurt, Germany \\
	\texttt{roig@cs.uni-frankfurt.de}
 \And
\href{https://orcid.org/0000-0002-0417-0336}{\includegraphics[scale=0.06]{orcid.pdf}\hspace{1mm}Bjoern M. Eskofier} \\
	Department AIBE, \\ FAU Erlangen-N\"urnberg \\ Erlangen, Germany and \\
 Institute of AI for Health, \\ Helmholtz Zentrum München, \\ Munich, Germany \\
	\texttt{bjoern.eskofier@fau.de}
  \And
\href{https://orcid.org/0000-0001-5886-0597}{\includegraphics[scale=0.06]{orcid.pdf}\hspace{1mm}Dario Zanca} \\
	Department AIBE\\
	FAU Erlangen-N\"urnberg\\
	Erlangen, Germany \\
	\texttt{dario.zanca@fau.de}
}

\date{}

\renewcommand{\headeright}{}
\renewcommand{\undertitle}{}
\renewcommand{\shorttitle}{FovEx}

\usepackage{eccvabbrv}

\usepackage{cleveref}

\begin{document}

\maketitle
\setlength{\tabcolsep}{4pt}

\begin{abstract}
Explainability in artificial intelligence (XAI) remains a crucial aspect for fostering trust and understanding in machine learning models. Current visual explanation techniques, such as gradient-based or class-activation-based methods, often exhibit a strong dependence on specific model architectures. Conversely, perturbation-based methods, despite being model-agnostic, are computationally expensive as they require evaluating models on a large number of forward passes. In this work, we introduce Foveation-based Explanations (FovEx), a novel XAI method inspired by human vision. FovEx seamlessly integrates biologically inspired perturbations by iteratively creating foveated renderings of the image and combines them with gradient-based visual explorations to determine locations of interest efficiently. These locations are selected to maximize the performance of the model to be explained with respect to the downstream task and then combined to generate an attribution map. We provide a thorough evaluation with qualitative and quantitative assessments on established benchmarks. Our method achieves state-of-the-art performance on both transformers (on 4 out of 5 metrics) and convolutional models (on 3 out of 5 metrics), demonstrating its versatility among various architectures. Furthermore, we show the alignment between the explanation map produced by FovEx and human gaze patterns (+14\% in NSS compared to RISE, +203\% in NSS compared to GradCAM). This comparison enhances our confidence in FovEx's ability to close the interpretation gap between humans and machines. \let\thefootnote\relax\footnotetext{This preprint has not undergone any post-submission improvements or corrections. The Version of Record of this article is published in International Journal of Computer Vision (Springer Nature), and is available online at \href{https://doi.org/10.1007/s11263-025-02543-y}{https://doi.org/10.1007/s11263-025-02543-y}}

\end{abstract}

\section{Introduction} \label{sec:intro}
In recent years, deep learning has made remarkable strides in revolutionizing computer vision, particularly in safety-critical domains such as medical imaging \cite{wang2022personalizing, ouyang2020self, fujiyoshi2019deep}, autonomous driving \cite{jing2022inaction, najibi2022motion}, industrial automation \cite{maschler2021deep, iqbal2019fault, liang2017using}, or security and surveillance \cite{singh2020real, xu2021deep, gruosso2021human}.
However, as these models become increasingly more complex, the lack of understanding of their decision-making processes poses significant challenges\cite{goodman2017european}. To address these challenges, there is a growing need for XAI methods aiming at ensuring transparency and interpretability to the black-box nature of deep learning models.

While a variety of explanation methods have been developed for vision models \cite{gradCAM, gradcampp, chefer2021transformer, chefer2021generic}, these approaches are often tailored to specific architectures and lack universality. GradCAM and its derivations \cite{gradCAM, gradcampp} have been originally described as effective class-specific XAI methods to compute gradient-weighted feature maps from the last layer of convolutional architectures, highlighting relevant regions in the input. Although the GradCAM method can be extended to vision transformers (through reshaping the feature maps and gradients from the deepest layers), the performance of this approach is adversely affected by certain architectural attributes of vision transformers, such as skip connections, non-local self-attention mechanisms, and unstable gradients \cite{Choi2023CVPR}.   
On the other side, XAI methods for vision transformers \cite{abnar2020quantifying, chefer2021transformer, chefer2021generic} are often tailored to transformer-specific characteristics, such as attention weights or class tokens, making their application to convolutional-based models unfeasible. Therefore, there is a pressing need for \textit{model-agnostic} XAI methods, i.e., an approach that can work on any model architecture without the need for changes or adaptations. This can ensure comparability in explanations across different architectures and ensure more reliable interpretations of deep neural networks. 


While current XAI methods provide insights into model decision-making processes, their quality often falls short when it comes to human understanding as they lack contextual aspects that make such explanations understandable to humans \cite{morrison2023evaluating, hsiao2023towards}.  Incorporating human-inspired constraints into XAI frameworks can enhance the quality of explanations and make them more aligned with human perception. One such fundamental aspect is foveated vision: humans' highest visual acuity occurs at the center of the visual field (fovea), while peripheral vision has a lower resolution, underlying the way humans prioritize details in a specific area. Recent work \cite{han2020scale, volokitin2017deep} has demonstrated the advantages of incorporating such constraints into vision models.

Deza \etal~\cite{deza2020emergent} show increased i.i.d. generalization as a computational consequence emerging with foveated processing. Location-dependent computation based on foveation have demonstrated efficiency and avoidance of spurious correlation from data, both for  convolutional~\cite{tiezzi2022foveated} and visual transformer~\cite{jonnalagadda2021foveater} models. Foveation priors are effective in generating visual scanpath~\cite{schwinn2022behind}. These results together demonstrate how introducing biological constraints in artificial neural networks both increases alignment with the human counterpart and fosters model performances. We believe that the aforementioned concepts and intuitions can open a promising novel avenue in the field of explainability.

In this paper, we address the challenges associated with current explainability approaches and propose a novel method, Foveation-based Explanations (FovEx), that is based on insights inspired by the biology of human vision. 
The final goal of \fvx{} is to determine explanations, in the form of attribution maps, on the output of a backbone model processing a given input pattern. 
To do so, input samples undergo a differentiable transformation mimicking human-foveated vision. Then, gradient information is leveraged to enable an iterative and post-hoc human-like exploration of the input image, to determine relevant image regions.

\begin{figure}[t!]
	\centering
	\includegraphics[scale=0.6]{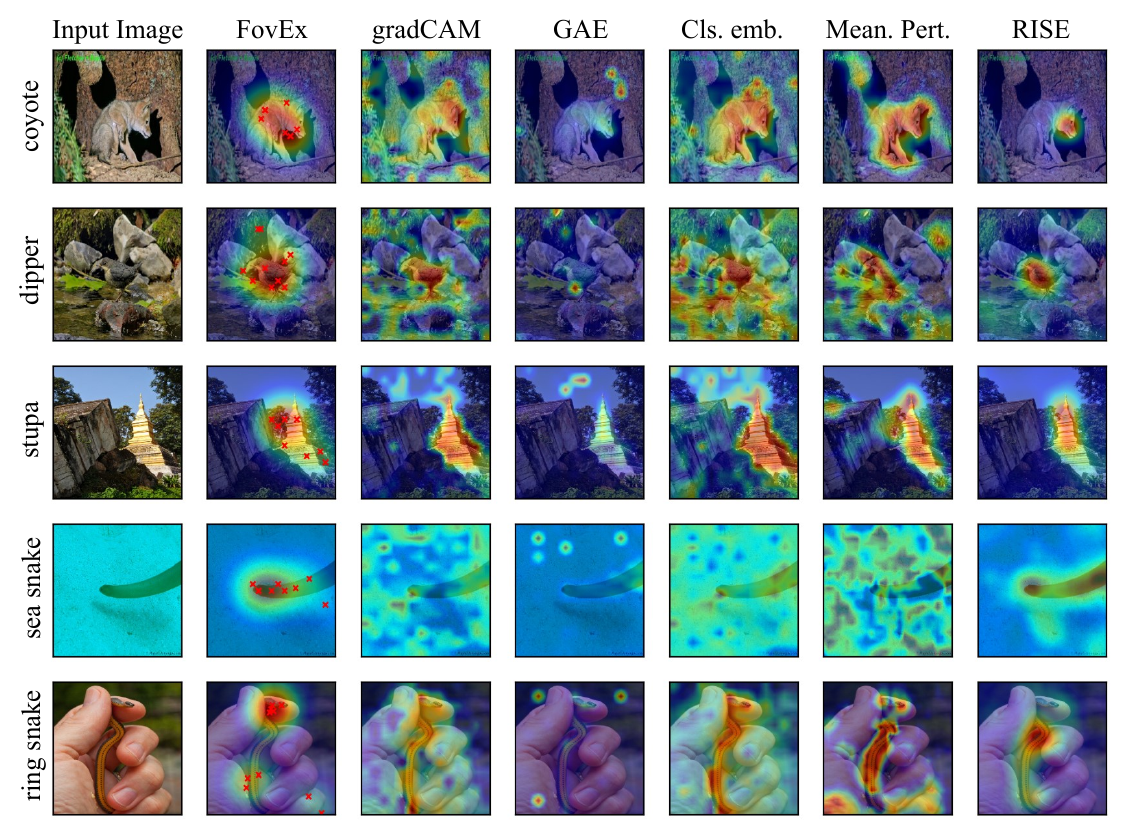}
	\caption{ViT-B/16 attribution maps. Explanation maps generated by FovEx (second column) and competitors for the ViT-B/16 predictor. Red crosses on explanation maps in the \fvx{} column denote fixation locations.}
	\label{fig:vit_viz}
\end{figure}

Such relevant regions are combined appropriately to generate attribution maps. The FovEx-generated explanation maps, compared in Figure \ref{fig:vit_viz} against state-of-the-art methods (further details in the remainder of the paper),  achieve state-of-the-art performance in common XAI metrics and have a better alignment with human gaze, as per experimental results.

To summarize, we delineate our contributions as follows:
\begin{itemize}
\item We introduce FovEx, a novel explanation method for DNNs that incorporates the biological constraint of human-foveated vision. FovEx extracts human-aligned visual explanations of the predictions in a post-hoc fashion, i.e., without introducing any architectural modification to the underlying predictor.
\item We demonstrate the effectiveness of FovEx through qualitative and quantitative evaluations for both convolutional and transformer-based models. We compare our approach to class-activation-based and perturbation-based XAI methods and demonstrate state-of-the-art performance for different model architectures.
\item We show that FovEx explanations enhance human interpretability via a quantitative investigation of the correlation between human gaze patterns and the explanation maps of DNNs generated by FovEx.
\end{itemize}

\section{Related Work} \label{sec:literature}
A variety of local post-hoc XAI methods exist for supervised image classification models \cite{ali2023explainable}. These techniques can be broadly categorized into ($i$) backpropagation or gradient-based methods, ($ii$) Class Activation Map (CAM)-based methods, and ($iii$) perturbation-based methods ~\cite{wagner2019interpretable, nguyen2023towards}. In this section, we first introduce the main literature for these three categories, and then provide concise insights into specific methods tailored for the interpretability of transformer-based models.

\noindent\textbf{Gradient-based Explanation Methods.} 
To generate explanations, gradient-based XAI approaches utilize the gradient of a pre-trained black-box model's output with respect to input features, i.e., image pixels~\cite{nielsen2022robust}. The seminal work by Simonyan \etal~\cite{Simonyan14a} constructs saliency maps by computing gradients of the non-linear class score function with respect to the input image. However, gradients frequently result in noisy visualizations. To address this issue,~\cite{smilkov2017smoothgrad} refines the explanation map by averaging over multiple saliency maps for a single input image. These saliency maps correspond to noisy duplicates of the input image, built by introducing random Gaussian noise with a zero mean into the original image. Layer-wise Relevance Propagation (LRP)~\cite{bach2015pixel} generates an explanation map by propagating fixed predefined decomposition rules from the output layer of a black-box model to the input layer. Softmax Gradient Layer-wise Relevance Propagation (SGLRP)~\cite{iwana2019explaining} utilizes the gradient of the softmax function to propagate decomposition rules, aiming to address the class-agnostic nature observed in vanilla LRP. Unlike gradient-based methods, our proposed approach uses gradient information exclusively to enable a human-like exploration of the input image to generate locations of interest.

\noindent\textbf{CAM-based Explanation Methods.} A class activation map (CAM)~\cite{zhou2016learning}  reveals the important areas in an image that a convolutional neural network relies on to recognize a particular class. CAM leverages the Global Average Pooling (GAP) layer and the top-most fully connected layer of convolution-based classification networks. 
Even if CAM produces class-discriminating saliency maps, it is fully dependant on architectural families and constraints. GradCAM~\cite{gradCAM} extends CAM by incorporating gradient information. It computes the gradient of the predicted class score with respect to the feature maps of the last convolutional layer. These gradients act as the weights of each feature map for the target class. GradCAM++~\cite{gradcampp} builds on GradCAM by employing pixel-wise weights rather than a single weight for a forward feature map of the final convolution layer. This enhancement enables GradCAM++ to preserve multiple instances of similar objects in the final explanation map. To further improve on the gradient-based CAM techniques,~\cite{wang2020score} uses class score as weights and~\cite{lee2021relevance} employs relevance score as weights in the explanation generation process. In contrast to CAM-based methods, our method is architecture agnostic, making it applicable to convolution and transformer-based models without any modifications.


\noindent\textbf{Perturbation-based Explanation Methods.} In the context of supervised image classification, perturbation-based explanation methods involve techniques that generate explanations by directly manipulating the pixels in the input image and observing the resultant changes in the output of the black-box model~\cite{ivanovs2021perturbation}. Various existing explanation techniques belong to this category~\cite{fong2017interpretable, Petsiuk2018rise, fong2019understanding, wagner2019interpretable}. Perturbation-based explanation methods are particularly versatile, holding the potential for application across various model architectures. However, these approaches exhibit high computational complexity as they rely on the computation of a large number of perturbations and model inference steps. Unlike common perturbation methods, our approach strategically determines optimal focus locations using gradient information, making it substantially more computationally efficient, see Section \ref{sec:comp}.

\noindent\textbf{Explaining Transformers.} While there is a variety of explanation methods available for convolution-based models, the range of methods for transformers is relatively limited. Chefer \etal \cite{chefer2021transformer} contribute to this domain by generating explanation maps for transformers with a method that combines LRP and gradient-based approaches.\cite{chefer2021generic} extends the application of the method proposed in \cite{chefer2021transformer} to provide explanations for any type of transformer model. Additionally, Vilas \etal \cite{vilas2023analyzing} propose an approach that quantifies how different regions of an image can contribute to producing a class representation in intermediate layers, using attention and gradient-based information. However, it is important to note that these methods cannot be seamlessly employed in a plug-and-play manner without making adjustments to pre-existing model implementations. Conversely, in this paper, we propose a novel method that can be applied to both convolution-based and transformer-based architectures without altering the neural architectures for computing the explanation.
\section{Foveation-based Explanation: \fvx{}} \label{sec:fovex}
Let us consider a black-box predictor $b(\cdot|\theta)$ defined for classification problems, that, without any loss in generality, we assume to be a neural network with learnable parameters $\theta$. The predictor is a function $b: \mathcal{X} \in \R^i \mapsto\mathcal{Y}$, that maps data instances $x$ from the input space $\mathcal{X} \in \R^i$ to the prediction $y$ in the target space $\mathcal{Y}$, containing the different labels to which the input data pattern can belong. We denote with $y=b(x|\theta)$ the prediction $y$ yielded by the predictor on the input pattern $x$.
We denote with $\theta_D$ the case in which the learnable parameters $\theta$ have been tuned on a training dataset $D$.

\begin{figure}[tb!]
	\centering
	\includegraphics[width=\linewidth]{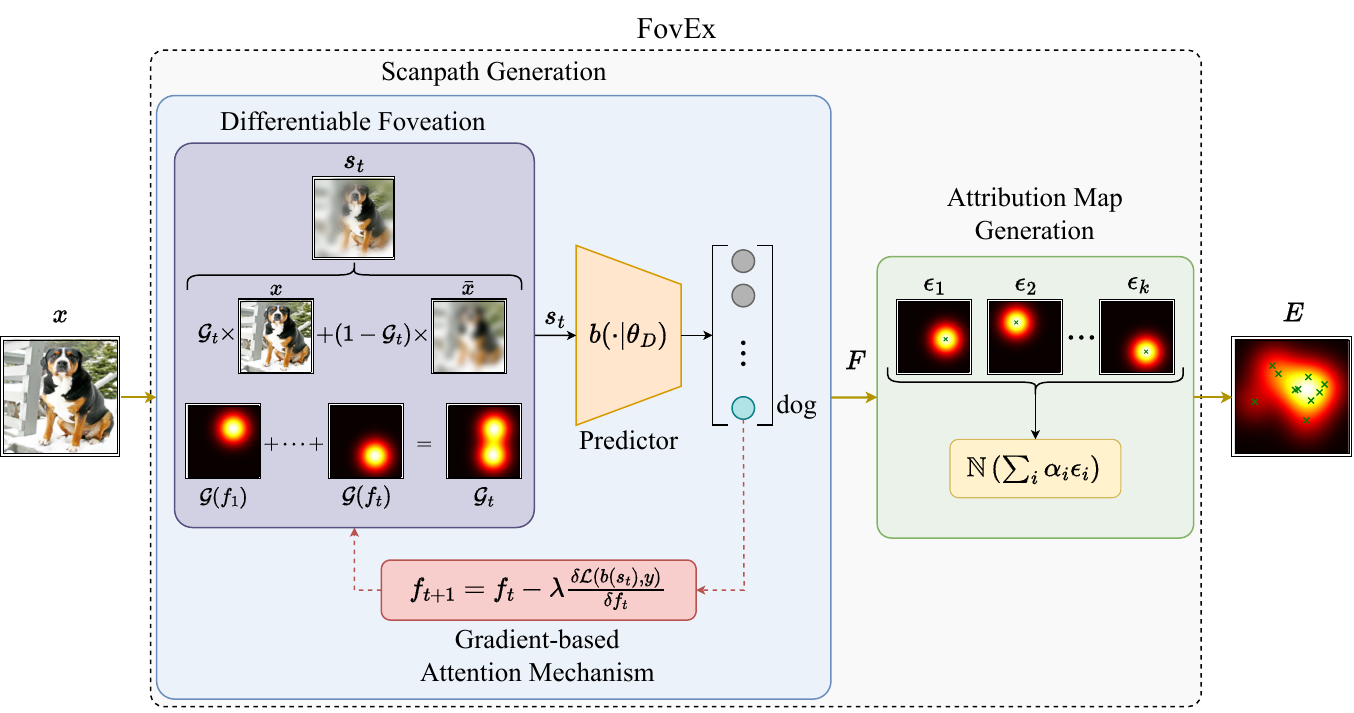}
	\caption{
                     The proposed \fvx{}. Given an input image $x$, \fvx{} produces an attribution map $E$ for a predictor $b$ trained on a dataset $D$. The image $x$ undergoes a differentiable foveation process yielding a transformed input ($s_t$) for the predictor $b$. The loss function $\mathcal{L}$, computed based on predicted class scores and ground truth $y$, is exploited by an attention mechanism for the generation of a sequence of fixations $(f_1, f_2, \cdots, f_t)$ (referred to as scanpath $F$). The resultant scanpath $F$ is employed to build a weighted linear combination of individual saliency maps $\epsilon_i$ associated with each fixation point $f_i$, leading to the final attribution map $E$.
                }
	\label{fig:FovEx}
\end{figure}

The goal of \fvx{} is to extract a human-understandable visual explanation of the decisions taken by the predictor $b$, in a post-hoc fashion, i.e., without introducing any architectural modification to the underlying predictor $b$.  Formally, \fvx{} is a function 
\begin{equation}
    E = \text{\fvx{}}(b, x)
    \label{fovex1st}
\end{equation}
where the output $E$ denotes the attribution map for the predictor $b$ associated to the input $x$. \fvx{} in turn consists of three fundamental operations, i.e., ($i$) a differentiable foveation mechanism, ($ii$) a gradient-based attention mechanism, and ($iii$) an attribution map generation process. A schematic illustration of the method is given in Figure \ref{fig:FovEx}. In the following, we give a formal definition of each operation.

\noindent\textbf{Differentiable Foveation}. 
Biological foveated vision is characterized by a central area of fine-grained processing  (i.e., the fovea) and a coarser peripheral area. We draw inspiration from Schwinn \etal\cite{schwinn2022behind} to design a differentiable foveation mechanism.
Let $x$ be an input image and $f_t$ be the current coordinates of the focus of attention at time $t$. First, we define a coarse version of $x$, denoted by $\bar{x}$, by convolving it with a Gaussian kernel, i.e., 
\begin{equation}
	\bar{x} = x * \mathcal{G}(0, \sigma_b^2)
	\label{eq:fmStep1}
\end{equation}
where $*$ represents the convolution operation, and $\mathcal{G}(0, \sigma_b^2)$ denotes the Gaussian kernel with a mean of $0$ and a standard deviation of $\sigma_b$. The parameter $\sigma_b$ controls the amount of blurring in the periphery of the image. From a biological standpoint, $\bar{x}$ can be regarded as the fundamental information obtained by peripheral vision within the initial milliseconds of stimulus presentation. A foveated input image $x_{\Phi}$ is obtained as a weighted sum of the original input $x$ and the coarse version of the input $\bar{x}$, i.e.,
\begin{equation}
	x_{\Phi} = \Phi \left(x, f_t \right) = \mathcal{W}(t) \cdot x + \left(1 - \mathcal{W}(t)\right) \cdot \bar{x}
	\label{eq:fmStep2}
\end{equation}
where $\cdot$ denotes the pixel-wise multiplication. In equation \ref{eq:fmStep2}, $\mathcal{W}(t) = \mathcal{G}\left(f_t, \sigma_f^2\right)$ stands for the pixel-wise weighting factor defined as the Gaussian blob $\mathcal{G}\left(f_t, \sigma_f^2\right)$ with a mean of $f_t$ and a standard deviation of $\sigma_f$, and $\Phi(\cdot)$ represents the foveation function. It is important to notice that, since all operations are differentiable, it makes the propagation of gradient information necessary for subsequent steps feasible. Additionally, the transformation described above introduces noise according to a foveal distribution $\mathcal{W}(t)$, perturbing the original input.

\noindent\textbf{Gradient-based Attention Mechanism}. The foveation mechanism allows for sequential exploration of the given input image. The next location of interest (i.e., the next fixation point) will depend on the state $s_t$ generated by all previous fixation locations, which can be regarded as the system's memory. The $s_t$ is obtained by cumulating Gaussian blobs to gradually expand the region of good visual fidelity after each fixation point, i.e.,
\begin{equation}
	s_t =  s(x, f_t) = \mathcal{G}_t \cdot x + (\vmathbb{1} - \mathcal{G}_t) \cdot \bar x 
	\label{inter_st}
\end{equation}
where $\mathcal{G}_t = \sum_{j=0}^{t}\beta^{j}\mathcal{G}_{t-j}\left(f_j, \sigma_f^2\right)$ symbolizes the cumulative Gaussian blob, whereas $\vmathbb{1}$ refers to a square matrix of ones. The forgetting factor $0 \le \beta \le 1$ regulates how much information is retained from previous fixations.

Let $\mathcal{L}\left(b(s_t), y\right)$ be the loss function at time $t$, e.g., calculated as the distance between the output predicted by $b$ for the current state $s_t$ and the target class $y_t$. The next fixation locations are dynamically adjusted to minimize the loss function $\mathcal{L}$ with respect to the current fixation location $f_t$, i.e., 
\begin{equation}
	f_{t+1} = f_t - \lambda \frac{\delta \mathcal{L}}{\delta f_t}
    \label{eq:gdOpt}
\end{equation}
where the hyperparameter $\lambda$ determines the step size at each iteration during optimization. The optimization technique iterates until a specified number of optimization steps  (\textsc{os}) have been performed, ensuring successful convergence. The influence of \textsc{os} is discussed in the ablation studies, in Section \ref{sec:ablation}. Random restarts (\textsc{rr}) are implemented when optimization fails to yield improvements in the loss function after a specified number of steps \textsc{os}, serving as a strategy to escape local minima. 
New fixations can be generated for an arbitrary number $N$ of steps, resulting in a sequence of $N$ fixations points, also called a \textit{scanpath}

\begin{equation}
	F = (f_1, ..., f_N)
\end{equation}

\noindent\textbf{Attribution Map Generation}. At this point, we have generated a sequence of $N$ fixation points based on an input and a task model. Each fixation point $f_i$ can be associated with a saliency map $\epsilon_i$, describing a 2D Gaussian distribution with a mean at $f_i$ and a standard deviation $\sigma_\epsilon$, where $\sigma_\epsilon$ is set to match the standard deviation of the Gaussian blob ($\sigma_f$). The final attribution map $E$, functioning as an explanation for the predictor $b$, is obtained as a weighted linear combination of the individual saliency maps associated with each fixation point, i.e., 
\begin{equation}
	E = \mathcal{N}\left(\sum_{i=1}^k 
        \alpha_i \epsilon_i\right)
	\label{eq:saliency}
\end{equation}
In Equation \ref{eq:saliency}, $\alpha_i$ denotes the weighting factor for saliency map $\epsilon_i$ and $\mathcal{N}(\cdot)$ represents min-max normalization. The weights $\alpha_i$ determine the contribution of each fixation to the final saliency map. In our experiments, we set $\alpha_i = 1$, $\forall I \in \{1, ..., N\}$, as different weighting schemes did not improve the quality of explanations on a validation set.
\section{Experiments} \label{sec:exsetup}

We conducted a comprehensive set of experiments to compare the performances of the proposed \fvx{} against state-of-the-art (SOTA) models and to showcase its ability to be agnostic to the architecture of the predictor $b$. We assessed \fvx{}'s performances in different scenarios, ranging from qualitative inspections and quantitative assessments in the common testbed of ImageNet-1K validation set \cite{ILSVRC15} to the correlation analysis of the generated attribution maps to human gaze. Additionally, we compared \fvx{}'s computational complexity against SOTA methods and performed in-depth model ablation studies. 
All our experiments were performed in a Linux environment, using an NVIDIA RTX 3080 GPU, and the implementation code can be found at \url{https://github.com/mahadev1995/FovEx}. 

\subsection{ImageNet-1K} \label{imagenetExp}

\textbf{Setup \& Data. } We selected two representative classification models as the predictor $b$ to be explained. In particular, we focused on a ResNet-50\footnote{\url{https://pytorch.org/vision/stable/models.html}} \cite{he2016deep} and a Vision Transformer (ViT-B/16)\footnote{\url{https://github.com/google-research/vision\_transformer}} \cite{dosovitskiy2020image}, that have been recently classified as foundation models \cite{bommasani2021opportunities}. The ResNet-50 model is pre-trained on the ImageNet-1K \cite{ILSVRC15} dataset, while the ViT-B/16 model is pre-trained on ImageNet-21K \cite{ridnik2021imagenet} dataset and fine-tuned on ImageNet-1K \cite{ILSVRC15} dataset. We assessed the model performances on a subset of 5000 images from the ImageNet-1K ~\cite{ILSVRC15} validation set, randomly sampled among the ones correctly classified by the predictor $b$ in order to measure the contribution of the method exactly, as pointed out by recent literature \cite{lee2021relevance}. The images are resized to a resolution of 224 $\times$ 224 pixels and normalized in the range of $(0,1)$. Performance on additional models is presented in Appendix \ref{secA1}. 

\noindent\textbf{Metrics. }  We report model performances focusing on the faithfulness and localization attributes of the explanation maps. Faithfulness measures the extent to which explanation maps correspond to the behavior of the black-box model. We report \textsc{Avg. \% drop} (lower is better) and \textsc{Avg. \% increase} introduced in ~\cite{gradcampp} (higher is better), as well as the \textsc{Delete} (lower is better) and \textsc{Insert} metrics (higher is better) proposed by Petsiuk \etal~\cite{Petsiuk2018rise}. 
The \textsc{Avg. \% drop} metric 
assesses the shift in confidence between two scenarios: one with the entire image as input and another with input limited to the regions highlighted by the explanation map. The \textsc{Avg. \% increase} metric quantifies instances across the dataset where the model's confidence rises when only the highlighted regions from the explanation map are considered. 
The \textsc{Delete} metric is used to evaluate the decrease in predicted class probability when removing pixels with decreasing importance, according to the explanation map. On the other hand, the \textsc{Insert} metric measures the increase in estimated likelihood when adding the essential pixels to the input, ordered from most to least important as indicated by the attribution map.
Localization measures an explanation map's capability to focus on a specific region of interest. Although good performance on localization may not imply a good explanation, it can provide interesting insights nonetheless because explaining a model's localization decisions can help identify patterns in the data the model has learned \cite{lucieri2020explaining}.  We report the Energy-Based Pointing Game (EBPG) metric~\cite{wang2020score} (higher is better). The EBPG metric calculates the energy of attribution maps within a predefined bounding box of the target class.

\begin{table}[t]
	\centering
    
	\caption{ViT-B/16 quantitative evaluation. Average metrics on the considered subset of ImageNet-1K validation dataset. The best-performing model is in bold, and the second-best is underlined.}

	\begin{tabular}{lcccccccc}
		\toprule
		\begin{tabular}[c]{@{}l@{}}Eval.\\ Name\end{tabular} &     FovEx   &  \begin{tabular}[c]{@{}c@{}}grad\\ CAM\end{tabular}  &    GAE  &    \begin{tabular}[c]{@{}c@{}}Cls.\\ Emb.\end{tabular}  &    \begin{tabular}[c]{@{}c@{}}Mean.\\ Pert.\end{tabular} &   RISE     & \begin{tabular}[c]{@{}c@{}}random\\ CAM\end{tabular} \\ \midrule 
		\textsc{Avg. \% drop}  ($\downarrow$)   &     \textbf{13.970}  &      40.057   &    86.207  &     34.862    &      29.753    &   \underline{15.673}  &  80.714   \\
		\textsc{Avg. \% increase} ($\uparrow$)  &     \textbf{30.389} &      11.469   &    0.799   &      13.329   &      20.549    &   \underline{22.189}   &  1.789    \\
		\textsc{Delete} ($\downarrow$)    &    0.240    &      \underline{0.157}    &    0.172   &      \textbf{0.155}    &      0.200     &   0.158    &  0.395     \\
		\textsc{Insert} ($\uparrow$)   &    \textbf{0.840}    &      \underline{0.818}    &    0.806   &      0.817    &      0.674     &   0.782    &  0.682      \\
		EBPG  ($\uparrow$)   &   \textbf{47.705}   &     41.667    &    39.812  &      39.350   &      40.646    &   \underline{42.633}  &  35.708    \\
		\bottomrule
	\end{tabular}

	\label{tab:quantVIT16}
\end{table}

\begin{table}[t]
	\centering
	\caption{ResNet-50 quantitative evaluation. Average metrics on the considered subset of ImageNet-1K validation dataset. The best-performing model is in bold, and the second-best is underlined.}
	\begin{tabular}{lcccccccc}
		\toprule
		\begin{tabular}[c]{@{}l@{}}Eval.\\ Name\end{tabular} &     FovEx   &  \begin{tabular}[c]{@{}c@{}}grad\\ CAM\end{tabular}  &    \begin{tabular}[c]{@{}c@{}}grad\\ CAM++\end{tabular}   &      \begin{tabular}[c]{@{}c@{}}Mean.\\ Pert.\end{tabular} &   RISE    & \begin{tabular}[c]{@{}c@{}}random\\ CAM\end{tabular} \\ \midrule
		\textsc{Avg. \% drop}  ($\downarrow$)     &     \textbf{11.780}  &      21.718 &    19.863  &     85.973        &      \underline{11.885}    &   61.317     \\
		\textsc{Avg. \% increase} ($\uparrow$)  &     \textbf{61.849} &      43.669   &    45.069  &      4.700        &      \underline{55.489}    &   16.729     \\
		\textsc{Delete} ($\downarrow$)    &    0.151           &      0.108   &    0.113   &      \textbf{0.082}&      \underline{0.100}     &   0.212      \\
		\textsc{Insert} ($\uparrow$)   &    \textbf{0.374}    &      0.368    &    0.361   &      0.280    &      \underline{0.372}     &   0.287      \\
		EBPG  ($\uparrow$)  &   46.977   &     \textbf{48.658}    &    \underline{47.412}  &      42.725   &     43.312     &   38.118    \\
		\bottomrule
	\end{tabular}
	\label{tab:quantRes50}
\end{table}

\noindent\textbf{Compared Models \& Architecture Details. } We compared the performances attained by \fvx{} against various SOTA XAI techniques. 
When considering the ViT-B/16 predictor, we report performances obtained by gradCAM~\cite{gradCAM}, Meaningful Perturbation (Mean. Pert.)~\cite{fong2017interpretable}, RISE~\cite{Petsiuk2018rise}, GAE~\cite{chefer2021generic}, and class embedding projection (Cls. Emb.)~\cite{vilas2023analyzing}. We remark that perturbation-based methods (Mean. Pert., RISE) have not been previously tested with Transformer architectures. 
In the case of a ResNet-50 predictor, we consider gradCAM, gradCAM++~\cite{gradcampp}, Mean. Pert., and RISE as competitors. 
In all the settings, we also report a baseline technique for sanity check, referred to as RandomCAM. In particular, randomCAM generates class activation maps for a randomly selected class regardless of the black-box model's prediction. 
For competitors, we utilized hyper-parameters following the respective original implementations. When considering \fvx{}, we adhere to the parameters suggested in Schwinn \etal\cite{schwinn2022behind}, which are biologically plausible, as they are designed to mimic human vision. The effect of different choices for the hyperparameters is discussed in our ablation study discussed in Section \ref{sec:ablation}.

\begin{figure}[t!]
	\centering
	\includegraphics[scale=0.6]{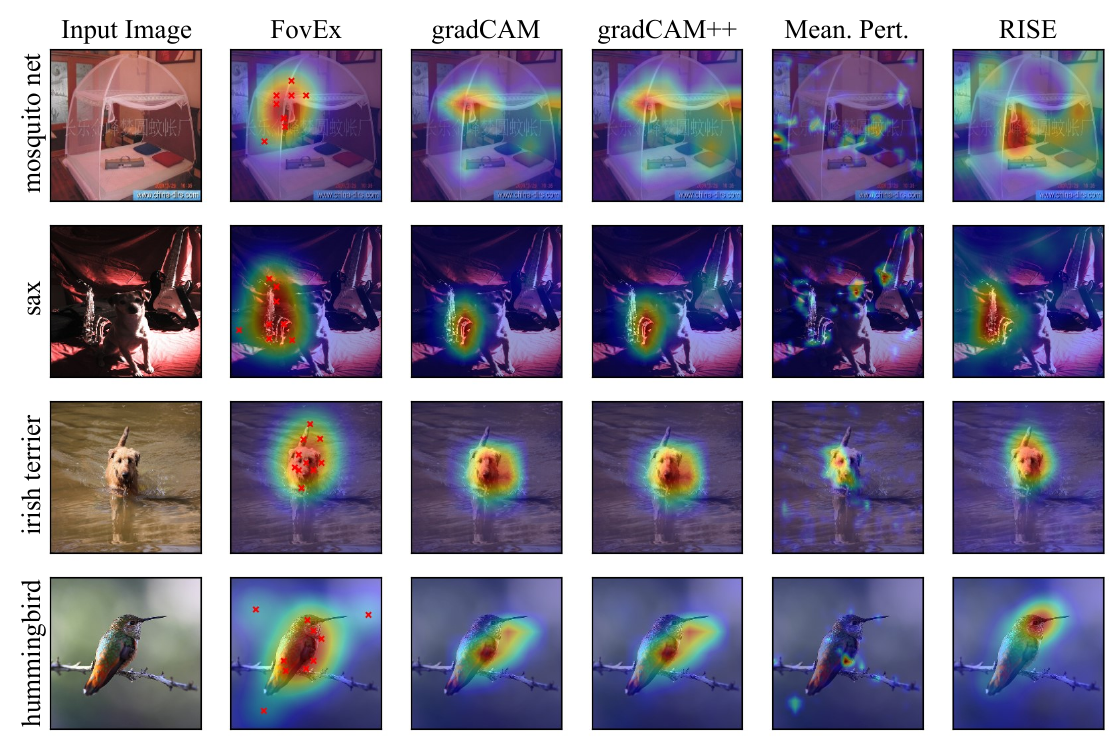}
	\caption{ResNet-50 attribution maps. Explanation maps generated by FovEx and competitors for the ResNet-50 model. Red crosses on explanation maps in the \fvx{} column denote fixation locations.}
	\label{fig:res_viz}
\end{figure}

\noindent\textbf{Quantitative Inspection. } We report in Table \ref{tab:quantVIT16} the results obtained for the ViT-B/16 predictor. 
\fvx{} outperforms all the competitors across all metrics, with the exception of  \textsc{Delete},  where the Cls. Emb. method excels. RISE secures the second-best position in \textsc{Avg. \% drop}, \textsc{Avg. \% increase}, and \textsc{EBPG} metrics, while gradCAM achieves the second-best performance in \textsc{Delete} and \textsc{Insert}. Similar conclusions can be drawn when we consider $b$ to be a ResNet-50, as reported in Table \ref{tab:quantRes50}. \fvx{} overcomes other methods in \textsc{Avg. \% drop}, \textsc{Avg. \% increase}, and \textsc{Insert}, while delivering competitive results in \textsc{Delete} and \textsc{EBPG}. 
We remark that while Mean. Pert. performs better than \fvx{} when considering \textsc{Delete}, it exhibits worse overall performances in all the other considered metrics. GradCAM is the top performer when considering \textsc{EBPG}, followed by gradCAM++. RISE secures the second position in all other metrics, except for EBPG.

These results showcase better faithfulness and localization capabilities of the proposed \fvx{} as compared to other methods, independent of the underlying model architecture. The poorer performances of \fvx{} on \textsc{Delete} can be attributed to the performed optimization scheme. Indeed, \fvx{} operates on a preservation ideology, where the optimization scheme aims to identify regions in the image that, when retained, allow the model to maintain its performance in correctly classifying input images. This design predisposes FovEx towards excelling in the \textsc{Insert} metric, which better aligns with \fvx{} optimization objective, but it exposes to downfalls when considering \textsc{Delete}. This is because the optimization process inherently focuses on finding the most critical regions for maintaining classification accuracy, rather than those whose removal would most degrade performance, leading to penalization in the \textsc{Delete} metric.

\noindent\textbf{Qualitative Inspection. } To better assess the explanations provided by the proposed \fvx{}, we conducted a qualitative evaluation comprising both a ($i$) visual comparison concerning the considered competitors and ($ii$) in-depth investigation regarding class-specific explanations. Figure \ref{fig:vit_viz} illustrates a comparison of attribution maps generated on several (rows) input samples (first column) by the proposed \fvx{} (second column) and the other considered competitors (subsequent columns), in the case of the ViT-B/16 predictor. \fvx{} generates explanation maps that are spread over the whole object of interest without any blemish or stray spot, such as the ones produced by gradCAM, GAE, and Cls. Emb. An explanation of such phenomenon is given by \cite{darcet2023vision}, demonstrating the presence of artifacts in the feature map of models, corresponding to high-norm tokens appearing during inference in unimportant background areas of images, repurposed for internal computations. Indeed, we remark how the issue is solved when using \fvx{} as the explanation method. Similarly, RISE  yields noise-free explanation maps, but we remark that its computational time is significantly higher compared to \fvx{} (see Section \ref{sec:comp}). 

\begin{figure}[t!]
	\centering
	\includegraphics[width=0.85\linewidth]{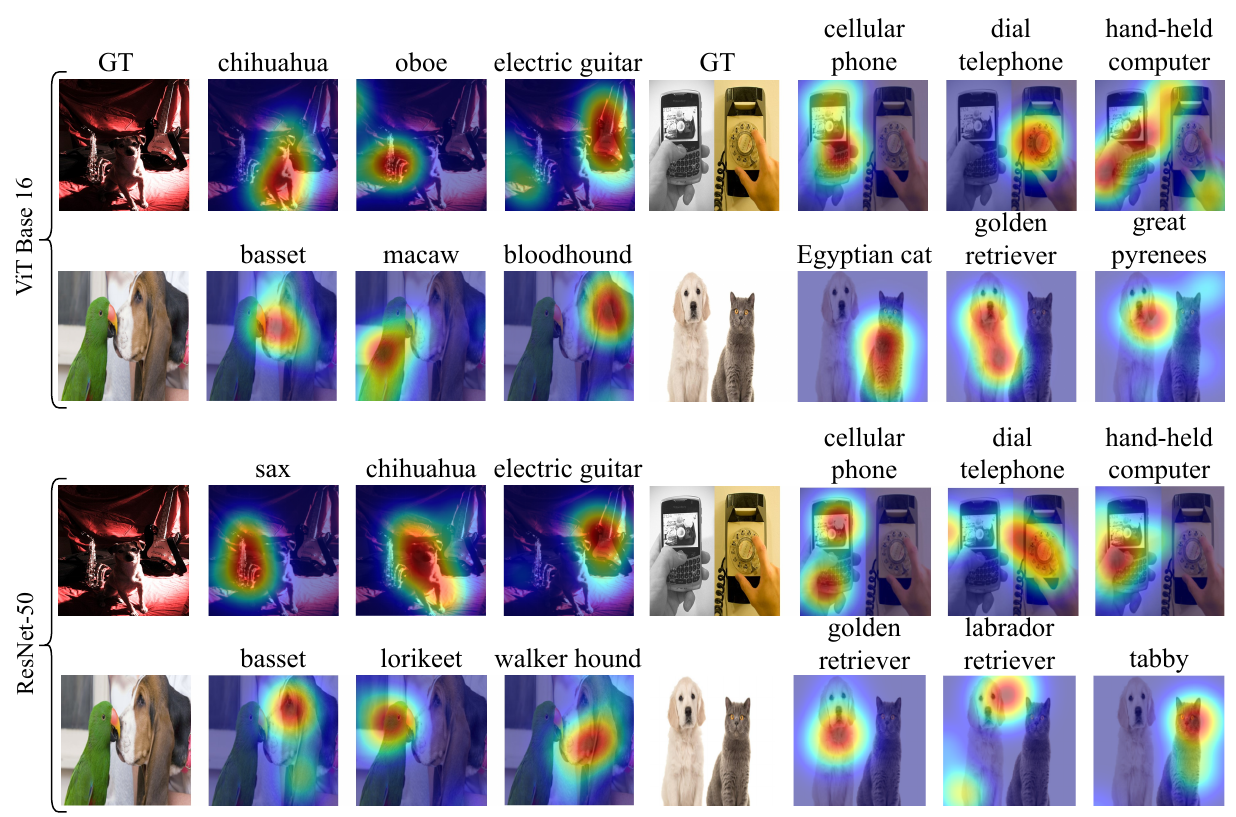}
	\caption{Class-discriminative evaluation. Class-specific attribution maps generated from \fvx{}, using ViT-B/16 (top) and ResNet-50 (bottom) as the predictor.}
	\label{fig:classSpViz}
\end{figure}

To better assess the quality of attribution maps generated by \fvx{} when dealing with convolution-based architectures, we report in Figure \ref{fig:res_viz} a visual comparison of attribution maps generated by FovEx (second column) and other SOTA methods for the ResNet-50 predictor. More visualizations of attribution maps for different models are given in Appendix \ref{secA3}. Also in this setting, \fvx{} is capable of generating attribution maps that are focused on the main object class contained in the processed image. The attribution map correctly spans over the object of interest,  thanks to the exploration carried on by the attention scanpath, differently from competitors (RISE, gradCAM) that tend to attribute the class prediction to certain object parts (bird head).

Additionally, we tested \fvx{}'s ability to localize class-discriminative information and capture fine-grained details. In particular, we selected images containing multiple annotated categories and we generated attribution maps for each of the available classes with \fvx{}. We report in Figure \ref{fig:classSpViz}-top the attribution maps obtained with ViT-B/16 as the predictor when generating explanations for different categories (indicated in the column header).  
Attribution maps generated from \fvx{} can discern the desired categories, localizing the relevant category inside the image with a good fine-grained definition of the pixels belonging to the object. Similar conclusions can be drawn when considering ResNet-50 as the predictor, as depicted in Figure \ref{fig:classSpViz}-bottom. 

We report the FovEx performance when considering three additional predictors ($b$) such as ConvNeXt \cite{liu2022convnet}, ViT-B/16 and ViT-B/32 from \texttt{torchvision}\footnote{Models from \texttt{torchvision} are trained on ImageNet-1K, while ViT-B/16 used in Section \ref{imagenetExp} is trained on ImageNet-21K and finetuned on ImageNet-1K. Source: \url{https://pytorch.org/vision/stable/models.html} } in the Appendix \ref{secA1}. 
In summary, FovEx continues to outperform other SOTA methods in the majority of the evaluation metrics considered for the aforementioned models.
\subsection{Human Gaze Correlation} \label{sec:evalGaze}
\textbf{Setup \& Data. } Generating attribution maps that are similar to the ones produced by human-gaze for image classification can enhance visual plausibility \cite{lai2020understanding, qi2023explanation}. Here,  we investigate the correlation of attribution maps generated by \fvx{} and other competitors with the ones obtained from human gaze for image free-viewing task. The goal is to investigate the possibility of generating explanation maps correlated to the general human gaze while being faithful to a black-box model's decision-making. We exploit the MIT1003 dataset~\cite{judd2009learning}, composed of 1003 images that are complemented by corresponding human attention maps,  collected in a controlled setting. 

\noindent\textbf{Metrics. } To quantitatively assess the similarity between the attribution maps generated by the black-box model and the human attention map, we employ the Normalised Scanpath Similarity (NSS) and Area Under the ROC Curve Judd version (AUCJ) metrics~\cite{zanca2018fixatons}. The higher the value of such metrics, the better.

\noindent\textbf{Compared Models \& Architecture Details. } We compare \fvx{} against the SOTA competitors described in the previous experiments. For this study, we focus on the ViT-B/16 predictor, pretrained on ImageNet-21K \cite{ridnik2021imagenet} and fine-tuned on ImageNet-1K \cite{ILSVRC15}. 


\begin{table}[tb]
	\centering
	\caption{Human-gaze correlation.  Quantitative evaluation results from human gaze experiment. We report NSS and AUCJ metrics (higher is better, see the main text for further details.)}
    \begin{tabular}{lcccccccc}
		\toprule
		\begin{tabular}[c]{@{}l@{}}Eval.\\ Name\end{tabular} &     FovEx   &  \begin{tabular}[c]{@{}c@{}}grad\\ CAM\end{tabular}  &    \begin{tabular}[c]{@{}c@{}}grad\\ CAM++\end{tabular}   & 
  Cls. Emb. & GAE & \begin{tabular}[c]{@{}c@{}}Mean.\\ Pert.\end{tabular} &   RISE     \\ \midrule
  NSS ($\uparrow$) & \textbf{0.7160} & 0.2357 & 0.0372 & 0.1231 & 0.4120 & 0.6197 & 0.6287 \\
  AUCJ ($\uparrow$) & \textbf{0.7044} & 0.5581 & 0.5094 & 0.5317 & 0.6875 & 0.6698 & 0.6400 \\
  \bottomrule
    \end{tabular}
	\label{tab:humanGazeViT}
\end{table}
\noindent\textbf{Results. } As reported in Table \ref{tab:humanGazeViT}, FovEx outperforms other methods in both the considered metrics. The achieved performances are almost doubled with respect to gradCAM and gradCAM++. Remarkably, explanation methods that have been proposed for the tested architecture (GAE, Cls. Emb.) are outperformed. 
This result highlights the alignment between the explanation maps produced by \fvx{} and human gaze patterns during free-viewing of natural images and enhances our confidence in \fvx{}'s ability to close the interpretation gap between humans and machines. Indeed, we remark that these results should be analyzed in conjunction with the ones about XAI metrics, summarised in Tables \ref{tab:quantRes50} and \ref{tab:quantVIT16}. Overall, to proposed \fvx{} is capable of coupling ($i$) the best alignment with human gaze patterns among the tested explanation methods, which we showed by investigating its correlation with human attention, together with ($ii$) extremely promising quantitative performances on several XAI metrics. This suggests that \fvx{} provides a more accurate and intuitive understanding of the model's decision-making process with respect to alternative methods, by allowing for a better comparison between human fixations and input localization importance of the model. 

\subsection{Efficiency Analysis} \label{sec:comp}
\begin{figure}[tb]
	\centering
	\includegraphics[width=0.85\linewidth]{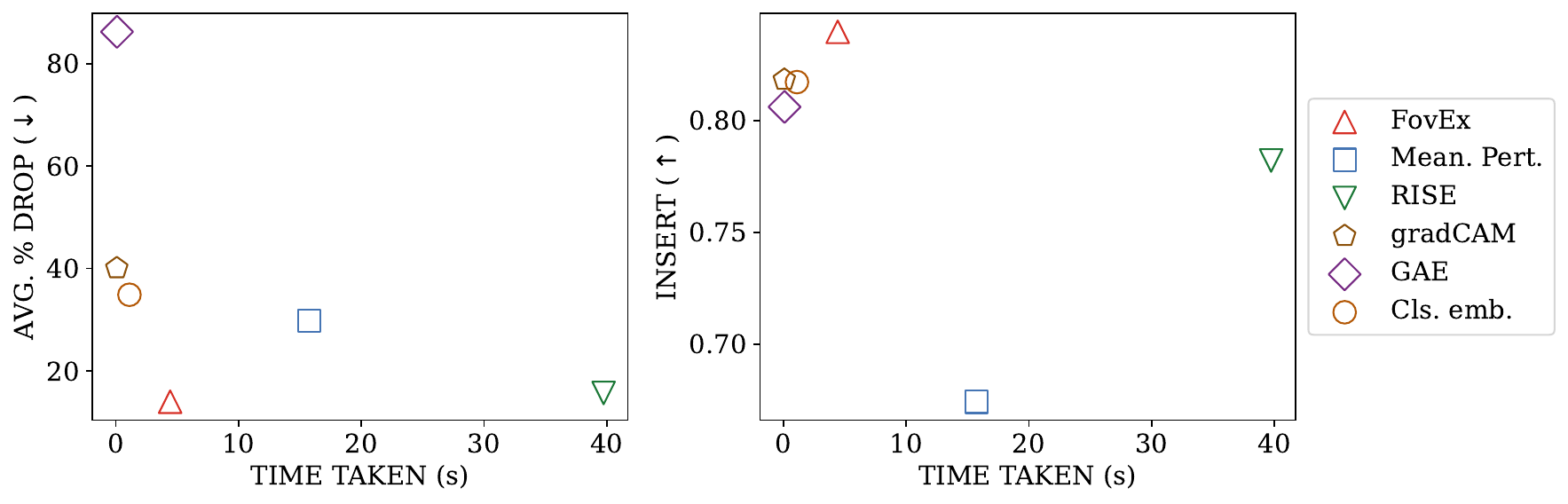}
	\caption{Efficiency analysis. Comparison of average attribution map generation time  ($x$-axis, seconds) against model performance ($y$-axis), both when considering \textsc{avg \% drop} (left) and \textsc{insert} metrics (right), for the ViT-B/16 model.}
	\label{fig:timeComplx}
\end{figure}
In this section, we investigate the time efficiency of the proposed \fvx{} compared with other SOTA approaches. We report the time necessary to produce an attribution map averaged over the considered 5000 images from the ImageNet-1K validation dataset. We focus our analysis on the ViT-B/16 predictor (a similar conclusion holds for the ResNet-50 architecture).  In Figure \ref{fig:timeComplx}-left we report the inference time ($x$-axis, seconds) against the \textsc{avg \% drop} metric ($y$-axis) for all the considered methods. The closer to the left-bottom corner, the better. Figure \ref{fig:timeComplx}-right depicts inference time ($x$-axis, seconds) against \textsc{Insert} ($y$-axis). In this case, the left-top corner implies better performance.
Overall, FovEx outperforms perturbation-based approaches (Mean. Pert. and RISE) in both performance and time efficiency. Conversely,  gradient-based methods are more time efficient but are far from the performances achieved by \fvx{} in the considered metrics. 

\subsection{Ablation studies}
\label{sec:ablation}
\begin{figure}[t!]
    \centering
    \includegraphics[width=0.9\linewidth]{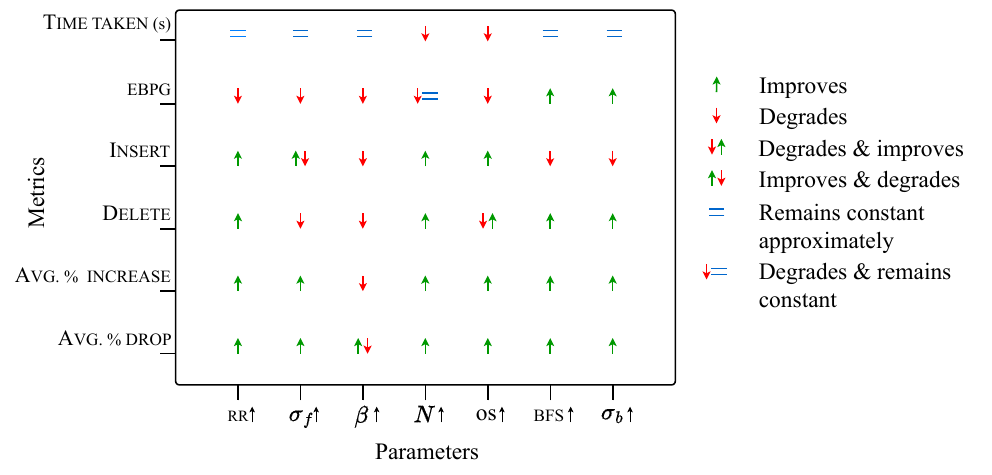}
    \caption{Summary of ablation study conducted to showcase the effect of various \fvx{} parameters. The $\uparrow$ near parameters ($x$ axis) denotes an increase in their value; except for rr, it signifies the value shifts from False to True.}
    \label{fig:ablationSummary}
\end{figure}

We undertook several ablation studies, which are delineated below. We conducted experiments to study the effect of various parameters such as random restart (\textsc{rr}), foveation sigma ($\sigma_f$), forgetting factor ($\beta$), scanpath length ($N$), optimization steps (\textsc{os}), blur sigma ($\sigma_b$), and blur filter size (\textsc{bfs}).  Figure \ref{fig:ablationSummary} depicts the summarized impact of the parameters mentioned above on the performance of \fvx{}. We utilized the ViT-B/16 model and the corresponding subset of the ImageNet-1K validation dataset employed in Section \ref{imagenetExp} for the ablation studies. Default values for the parameters are given in Appendix \ref{secA2}.

\begin{table}[t!]
	\centering
	\caption{Quantitative assessment. Average metrics on the considered subset of ImageNet-1K validation dataset for ViT-B/16 model for different \textsc{rr} values. Bold terms denote the best performance and underlined terms represent the second best performance.}
	\begin{tabular}{lccccc}
		\toprule
		\begin{tabular}[c]{@{}c@{}}Eval.\\ Name\end{tabular} &  \begin{tabular}[c]{@{}c@{}}\textsc{Avg. \% drop}\\ ($\downarrow$) \end{tabular} &  \begin{tabular}[c]{@{}c@{}}\textsc{Avg. \% increase}\\ ($\uparrow$) \end{tabular} & \textsc{Delete} ($\downarrow$)  & \textsc{Insert} ($\uparrow$) & \textsc{EBPG}  ($\uparrow$) \\ \midrule 
		\textsc{rr} = True  & \textbf{13.973} & \textbf{30.389} & \textbf{0.240} & \textbf{0.840} & \underline{47.705} \\
		\textsc{rr} = False & \underline{17.561} & \underline{26.389} & \underline{0.247} & \underline{0.824} & \textbf{51.067}    \\
		\bottomrule
	\end{tabular}
	\label{tab:rrQuant}
\end{table}

\begin{figure}[tb]
	\centering
	\includegraphics[width=0.65\linewidth]{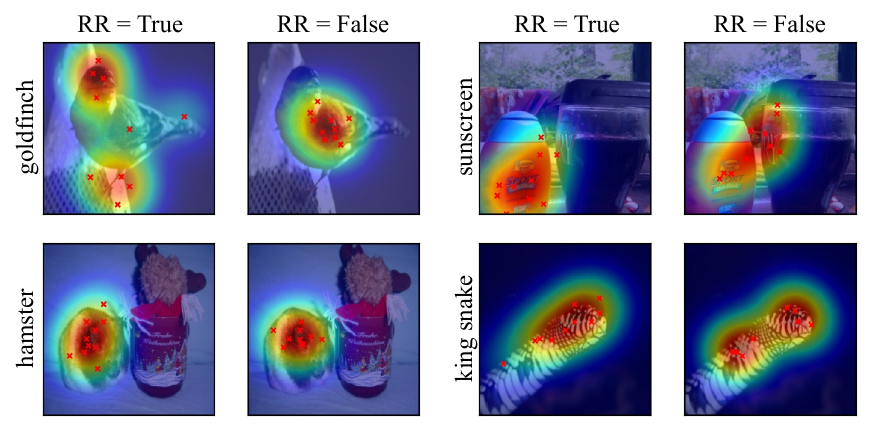}
	\caption{Qualitative assessment. Comparison of attribution maps generated using FovEx with different \textsc{rr} values.}
	\label{fig:randomRestart}
\end{figure}

\noindent\textbf{Random Restart} (\textsc{rr}). 
The \textsc{rr} parameter helps to avoid local minima in the optimization scheme for finding the location of fixation $f_t$. When \textsc{rr} is set to True, the exploration starts at a completely random location. Figure \ref{fig:randomRestart} depicts the visual comparison of explanation maps generated using two different values of \textsc{rr}. On close inspection, it is evident that fixation locations are closer to each other when \textsc{rr} is set to False. Table \ref{tab:rrQuant} presents the comparison of quantitative evaluation for different configurations of \textsc{rr}. With \textsc{rr} set to True, we get better performance in every metric except for the \textsc{EBPG} metric. 
\begin{figure}[t!]
	\centering
	\includegraphics[width=0.65\linewidth]{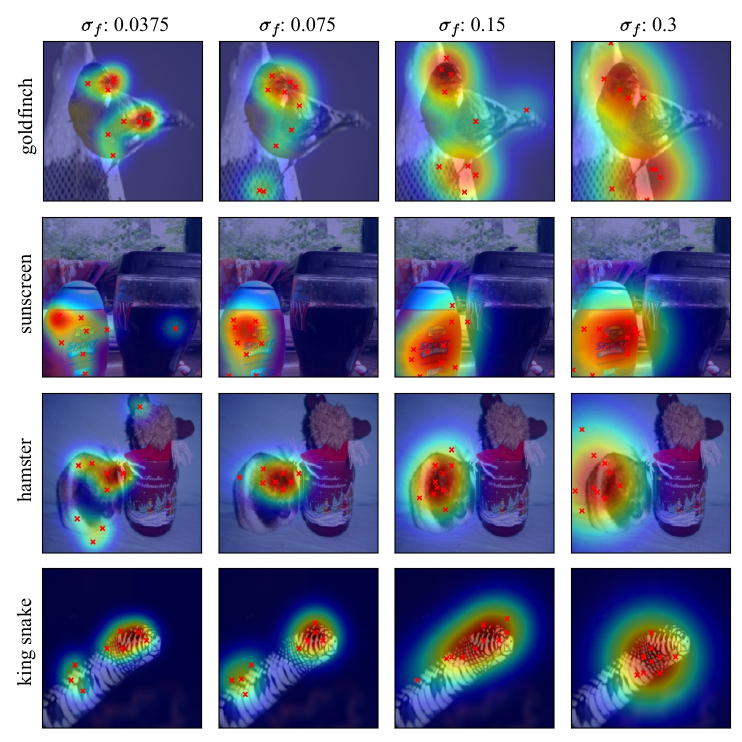}
	\caption{Qualitative assessment. Comparison of attribution maps generated using FovEx
with different $\sigma_f$ values.}
	\label{fovsig}
\end{figure}

\noindent\textbf{Foveation Sigma ($\sigma_f$). }
The $\sigma_f$ parameter controls the standard deviation of the area with higher visual acuity in the foveated input image. A higher value of $\sigma_f$ results in a larger region with enhanced visual clarity. In our experiments, we create attribution maps with varying $\sigma_f$ values- specifically, \{0.3, 0.15, 0.075, 0.0375\}.

\begin{figure}[t!]
	\centering
	\includegraphics[width=0.9\linewidth]{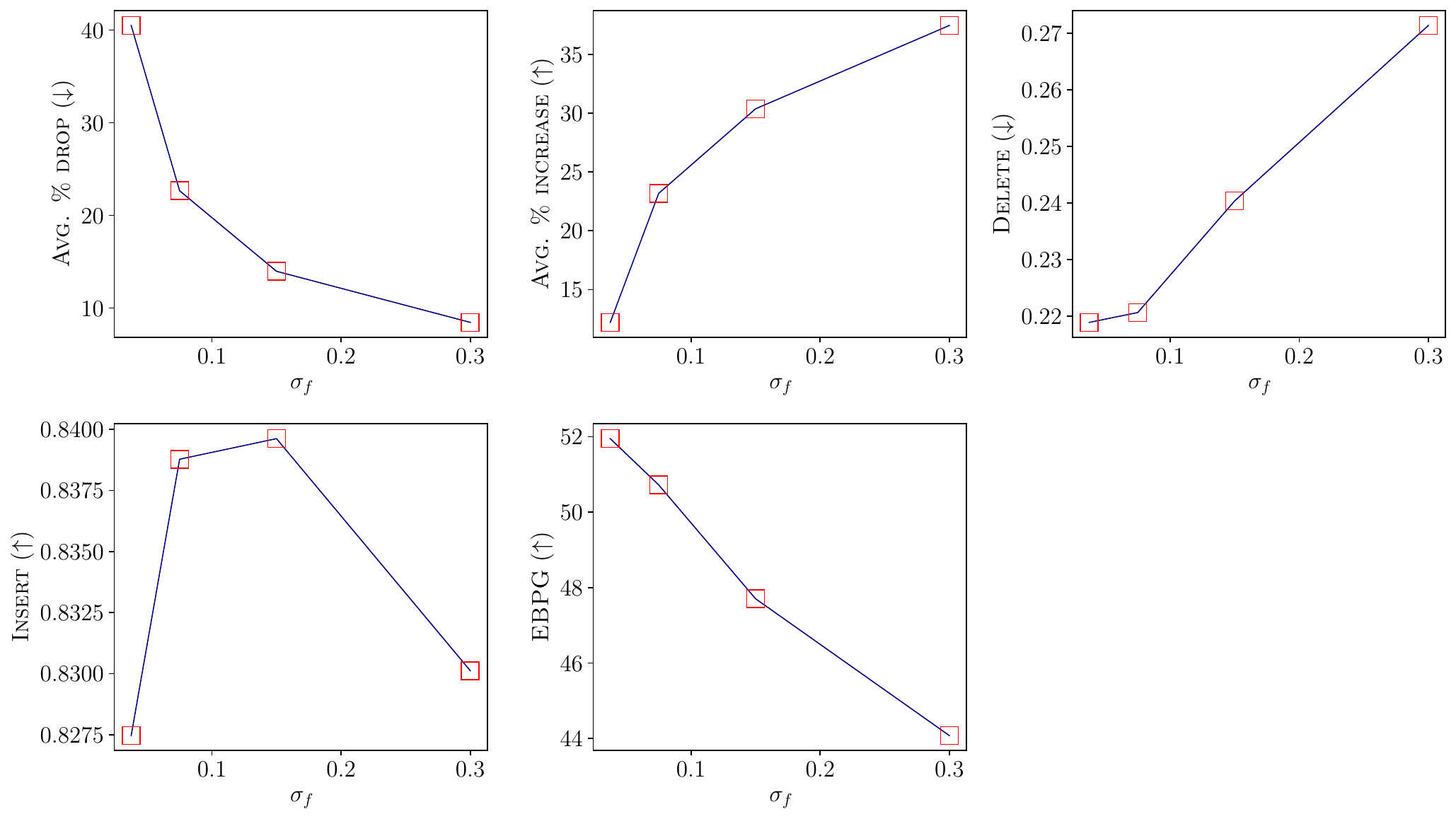}
	\caption{Quantitative assessment. Illustration of variation in performance of FovEx with respect to $\sigma_f$.}
	\label{FovSigma}
\end{figure}
\begin{figure}[t!]
    \centering
    \includegraphics[width=0.9\linewidth]{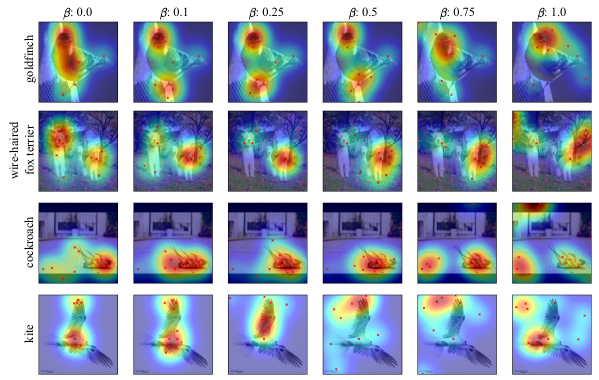}
    \caption{Qualitative assessment. Comparison of attribution maps generated using FovEx with different $\beta$ values.}
    \label{fig:betaQual}
\end{figure}
The visual comparison shown in Figure \ref{fovsig} indicates an increase in $\sigma_f$ values results in attribution maps with larger areas, while lower $\sigma_f$ values result in a more concentrated explanation map. We report quantitative assessments in Figure \ref{FovSigma}. As the value of $\sigma_f$ increases, there is an improvement in the metrics of \textsc{Avg. \% drop} and \textsc{Avg. \% increase}. However, the metric of \textsc{Delete} decreases. It is worth noting that an optimal $\sigma_f$ value of 0.15 shows the maximum performance in the \textsc{Insert} metric. Increasing $\sigma_f$ leads to a wider spread of the attribution map, compromising localization. 

\begin{figure}[t!]
	\centering
	\includegraphics[width=\linewidth]{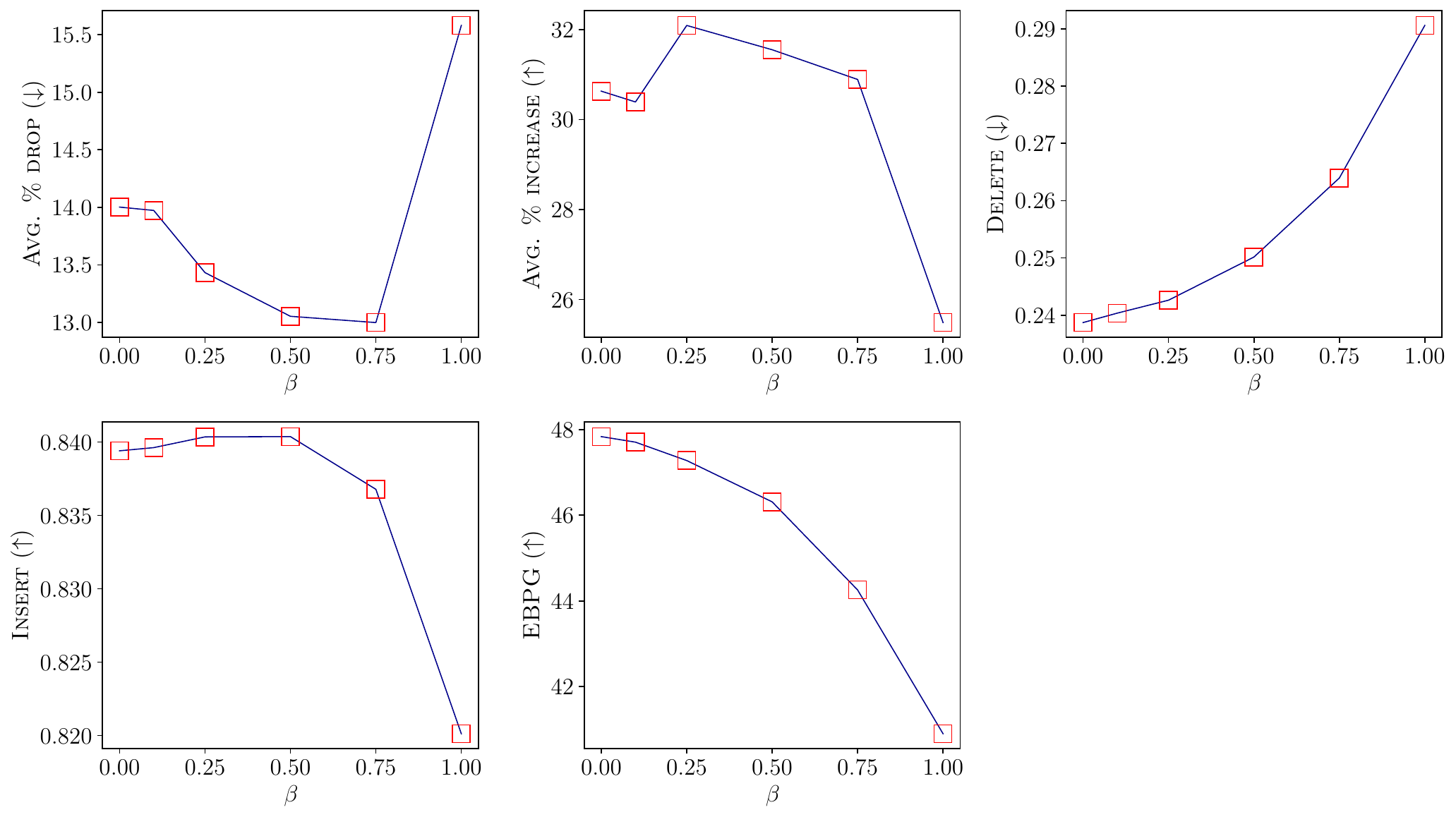}
	\caption{Quantitative assessment. Illustration of variation in performance of FovEx with respect to $\beta$.}
	\label{fig:betaQuant}
\end{figure}

\noindent\textbf{Forgetting Factor ($\beta$). } 
The forgetting factor $0 \le \beta \ge 1$ regulates how much information is retained from previous fixations. In our study, we created attribution maps for various $\beta$ values to investigate their effects. Specifically, we used the values \{0.0, 0.1, 0.25, 0.5, 0.75, 1.0\}. Visual comparison illustrated in Figure \ref{fig:betaQual}  reveals that the quality of explanation maps decreases with an increase in the $\beta$ value. However, the quantitative evaluation paints a nuanced picture as depicted in Figure \ref{fig:betaQuant}. 
\begin{figure}[t!]
    \centering
    \includegraphics[width=0.9\linewidth]{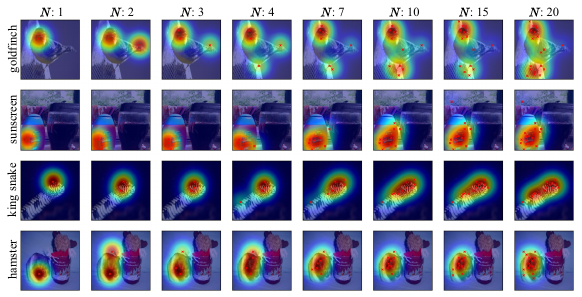}
    \caption{Qualitative assessment. Visual comparison of attribution maps generated using FovEx with different $N$ values.}
    \label{fig:QualN}
\end{figure}

\begin{figure}[t!]
	\centering
	\includegraphics[width=0.9\linewidth]{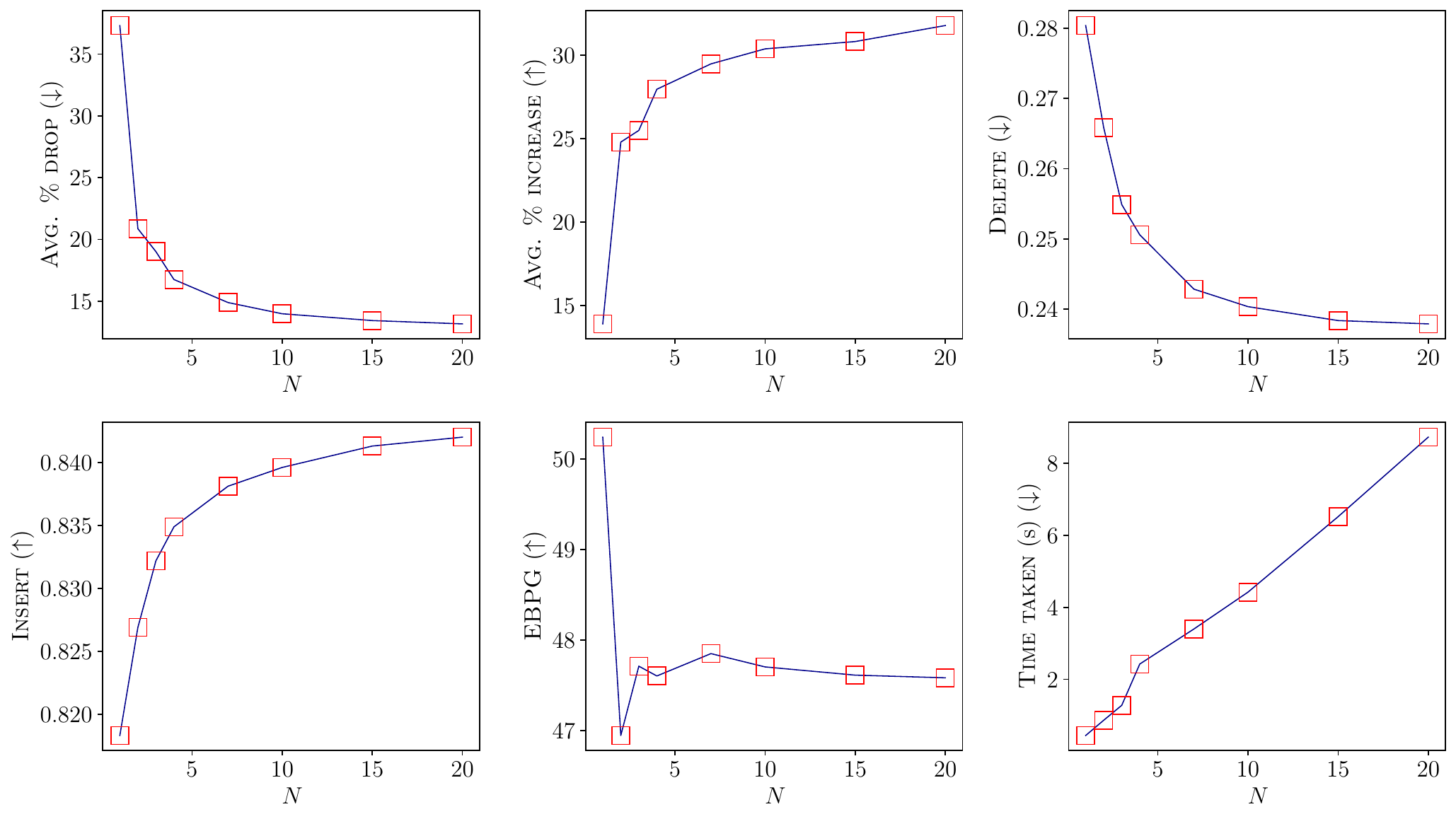}
	\caption{Quantitative assessment. Illustration of variation in performance of FovEx with respect to $N$.}
	\label{fig:QuantN}
\end{figure}

\noindent\textbf{Scanpath Length ($N$). }
$N$ controls the length of the scanpath $F$, i.e., the number of fixation points computed to generate an attribution map $E$. To study the impact of $N$ on the performance of FovEx, we generate attribution maps considering the following values for $N$, \{1, 2, 3, 4, 7, 10, 15, 20\}. As depicted in Figure \ref{fig:QualN} and Figure \ref{fig:QuantN}, both qualitative and quantitative (except \textsc{EBPG} and \textsc{Time taken}) performance improve with an increase in the value of $N$. However, the average time taken to produce an attribution map increases monotonically with $N$. 
\begin{figure}[t!]
    \centering
    \includegraphics[width=0.8\linewidth]{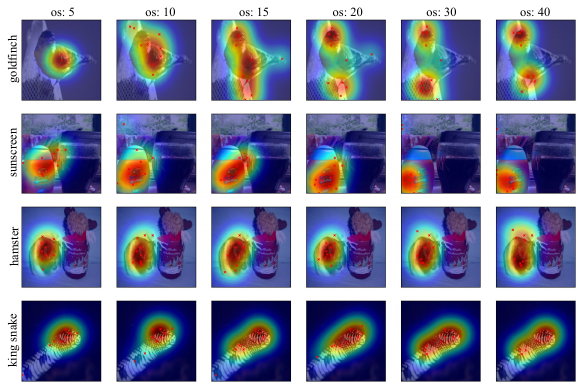}
    \caption{Qualitative assessment. Visual comparison of attribution maps generated using FovEx with different \textsc{os} values.}
    \label{fig:Qualos}
\end{figure}

\begin{figure}[t!]
	\centering
	\includegraphics[width=0.9\linewidth]{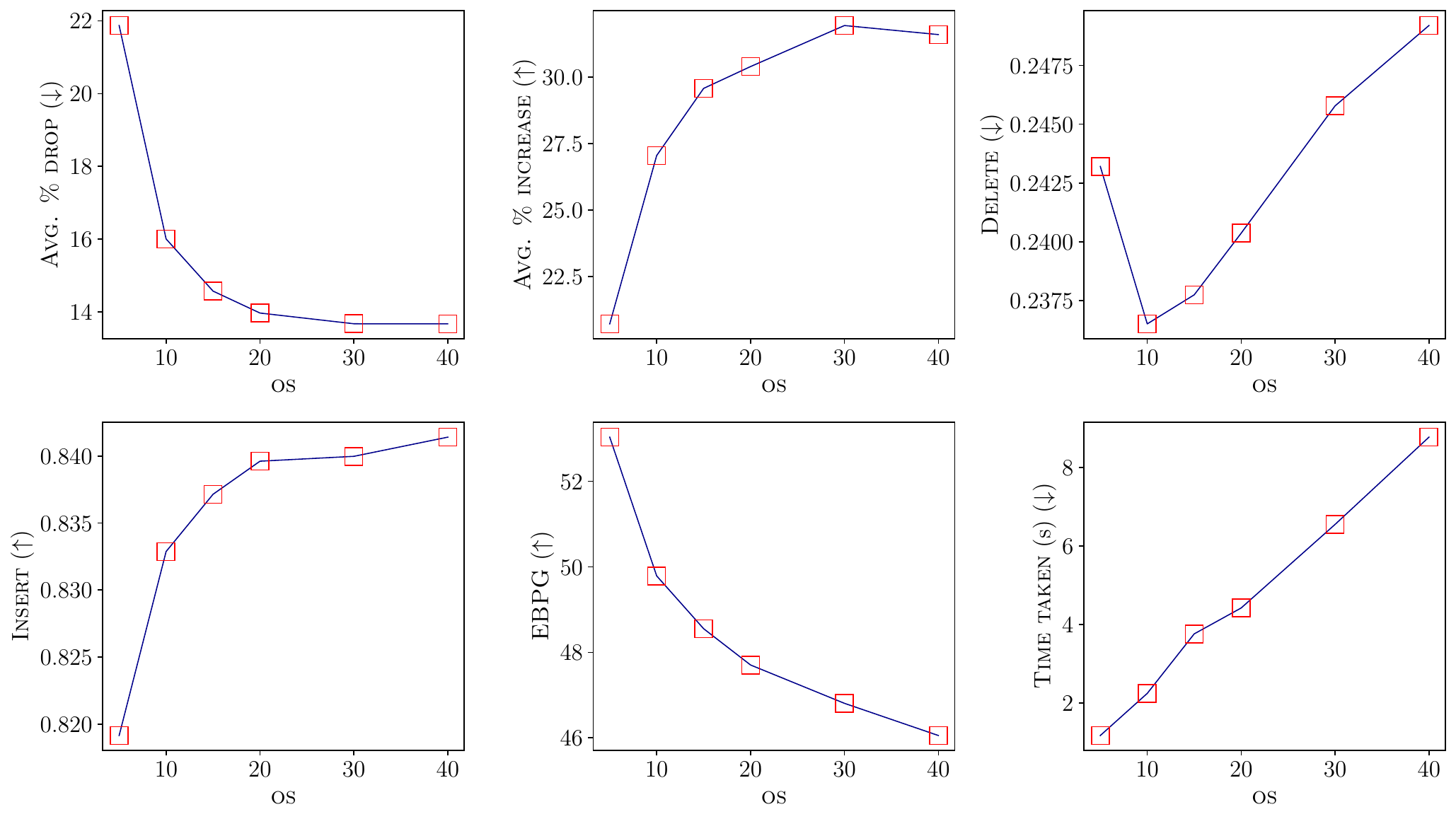}
	\caption{Quantitative assessment. Illustration of variation in performance of FovEx with respect to \textsc{os}.}
	\label{fig:Quantos}
\end{figure}

\noindent\textbf{Optimization Steps} (\textsc{os}). 
The \textsc{os} parameter controls the number of iterations performed in the optimization scheme to find a fixation location. We generate attribution maps using the following values of \textsc{os}, \{5, 10, 15, 20, 30, 40\}. The performance of FovEx improves with increment in \textsc{os} values both qualitatively and quantitatively (except for \textsc{Delete} and \textsc{EBPG} metric) as illustrated in Figure \ref{fig:Qualos} and Figure \ref{fig:Quantos}. Similar to the effect of $N$ with an increase in \textsc{os}, the average time taken to produce an attribution map increases.
\begin{figure}[t!]
    \centering
    \includegraphics[width=0.55\linewidth]{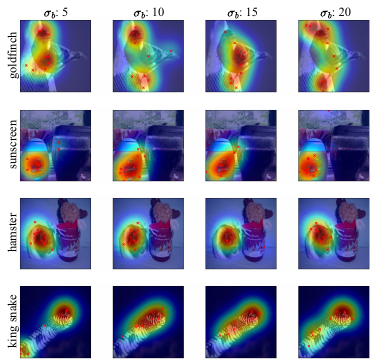}
    \caption{Qualitative assessment. Visual comparison of attribution maps generated using FovEx with different $\sigma_b$ values.}
    \label{fig:QualSigmab}
\end{figure}

\begin{figure}[t!]
	\centering
	\includegraphics[width=0.9\linewidth]{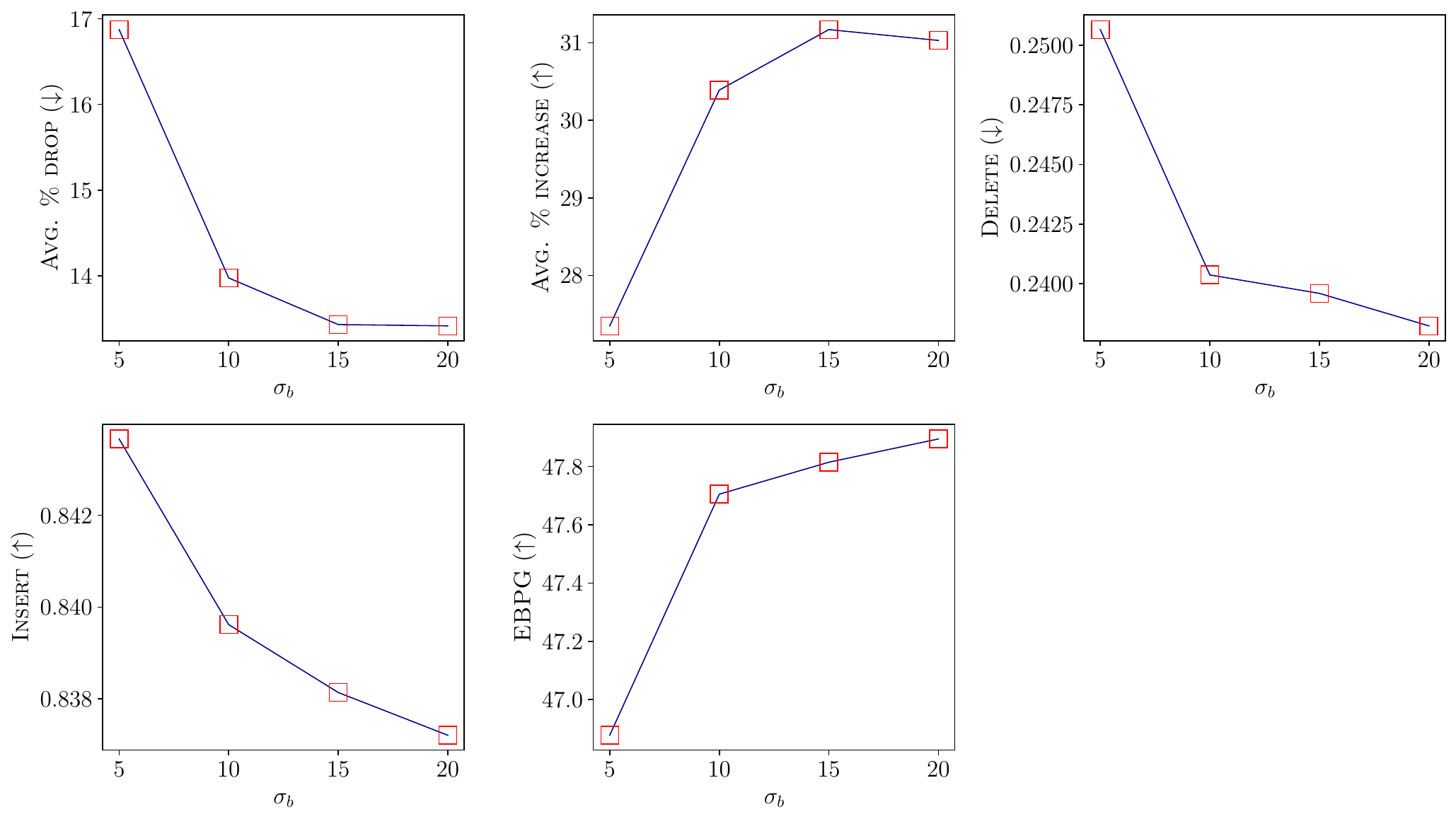}
	\caption{Quantitative assessment. Illustration of variation in performance of FovEx with respect to $\sigma_b$.}
	\label{fig:QuantSigmab}
\end{figure}

\noindent\textbf{Blur Sigma ($\sigma_b$). } The $\sigma_b$ parameter represents the standard deviation of the filter values used to create a blurred version of the input image which constitutes the peripheral region of the final foveated input image. We consider the following values for $\sigma_b$, \{5, 10, 15, 20\} for the experiments. Visual comparison in Figure \ref{fig:QualSigmab} depicts plausible attribution maps generated at all $\sigma_b$ values. However, the quantitative assessment illustrated in Figure \ref{fig:QuantSigmab} showcases an improvement in performance with increment in the value of $\sigma_b$ except in the \textsc{Insert} metric.
\begin{figure}[t!]
    \centering
    \includegraphics[width=0.65\linewidth]{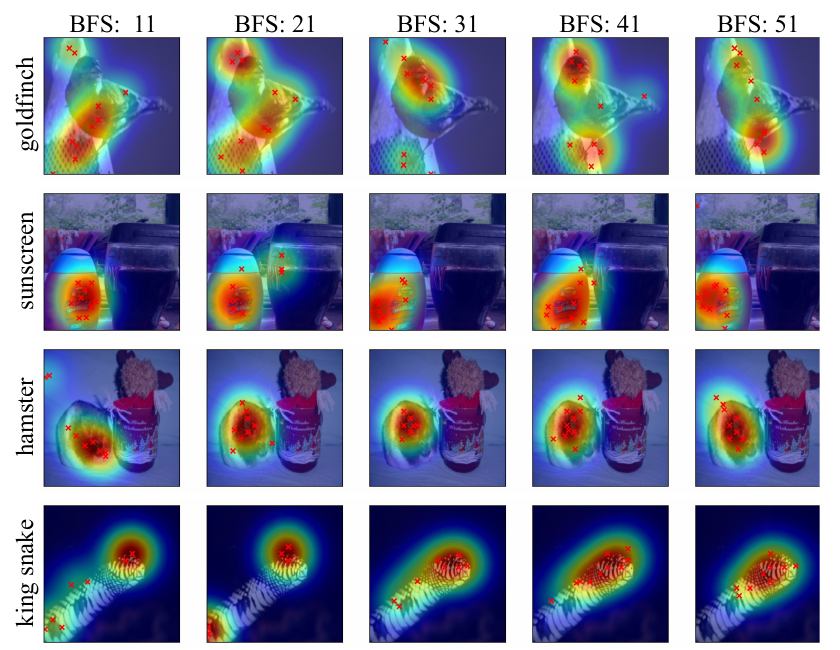}
    \caption{Qualitative assessment. Visual comparison of attribution maps generated using FovEx with different \textsc{bfs} values.}
    \label{fig:Qualbfs}
\end{figure}

\begin{figure}[t!]
	\centering
	\includegraphics[width=0.9\linewidth]{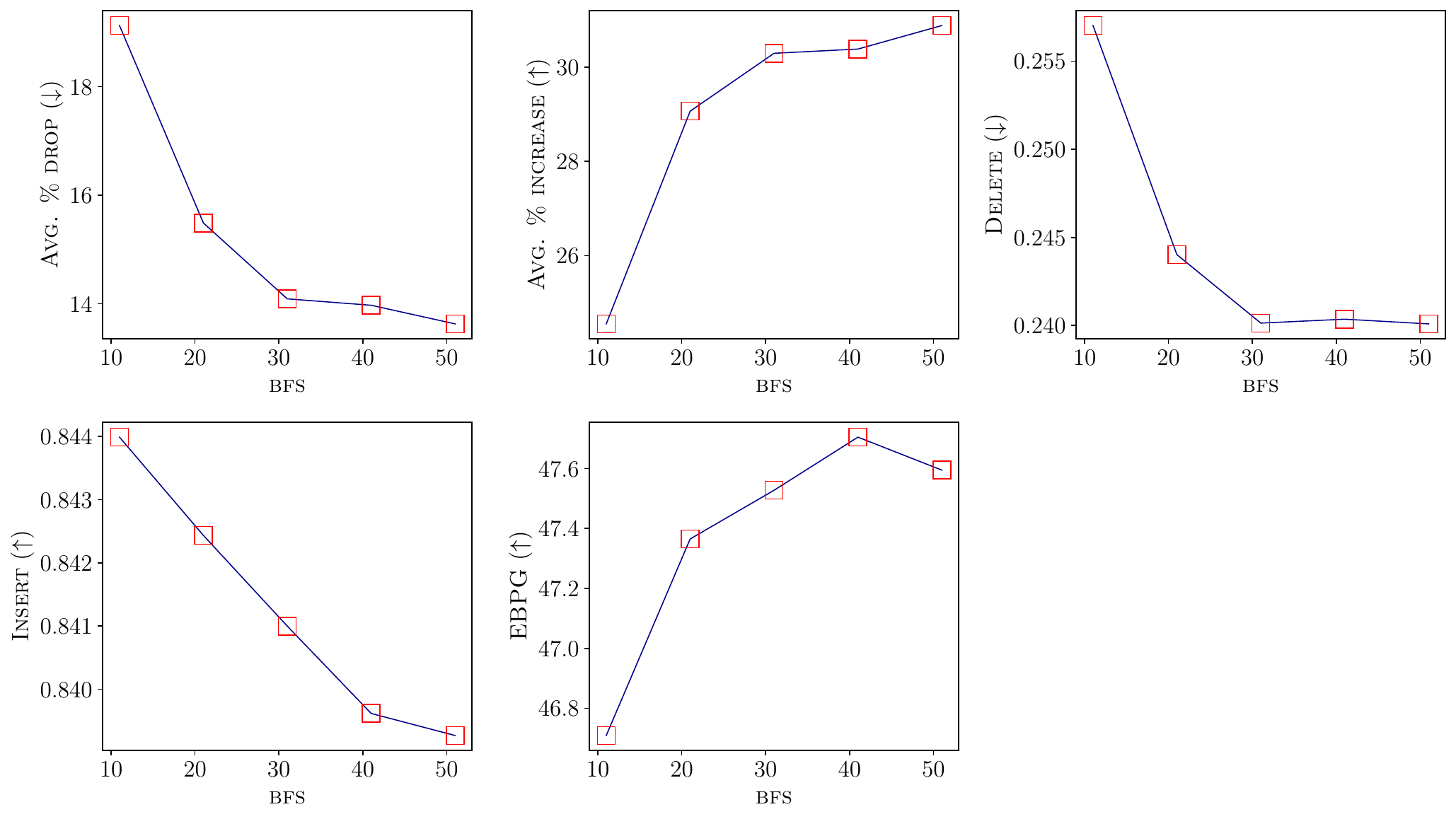}
	\caption{Quantitative assessment. Illustration of variation in performance of FovEx with respect to \textsc{bfs}.}
	\label{fig:Quantbfs}
\end{figure}

\noindent\textbf{Blur Filter Size} (\textsc{bfs}). 
The \textsc{bfs} parameter represents the size of the Gaussian kernel used to create the blurred version of the input image. To study the effect \textsc{bfs} we consider the following values, \{11, 21, 31, 41, 51\}. Qualitative assessment illustrated in Figure \ref{fig:Qualbfs}, showcases visually slightly better attribution maps at \textsc{bfs} value of 41 than other values. Figure \ref{fig:Quantbfs} illustrates the quantitative assessment. With an increase in \textsc{bfs} value the performance of FovEx increases except in \textsc{Insert} metric. 
\section{Conclusion} \label{sec:conclusion}
We introduced FovEx, a novel explanation technique that integrates foveated human vision principles into the explanation generation process for black-box models. FovEx not only generates visually coherent explanation maps for both transformer and convolution-based models but also possesses the ability to discern between different classes. It demonstrates superior performance in quantitative evaluations, outperforming competing methods for 4 out of 5 evaluation metrics for the ViT-B/16 model and for 3 out of 5 metrics for ResNet-50. Furthermore, incorporating biological foveated vision concepts allows FovEx to generate explanation maps that have a higher correlation to human gaze patterns on the considered dataset, as illustrated in Section \ref{sec:evalGaze}, paving the way for a more accurate and intuitive understanding of model's decisions. 

FovEx surpasses gradient-based and perturbation-based methods in all faithfulness metrics except the \textsc{Delete} metric, for which it falls short in comparison to gradient-based methods. This is mostly due to FovEx's optimization scheme prioritizing the selection of image regions that are the most informative for the explanation, thereby offering a specific advantage in classification performance preservation, as reflected by other metrics such as \textsc{Insert}.
From an ethical perspective, we notice that the use of a small, potentially non-representative human gaze dataset might raise biases, and results on human alignment could not be representative of the diversity of a larger population. However, our method opens the door to better comparing model and human gaze pattern behavior. 

Future work will extend the applicability of FovEx to models in various task domains, such as visual question answering and image captioning.

\bibliographystyle{unsrtnat}
\bibliography{egbib}


\begin{appendices}
\section{Performance on Various Models}\label{secA1}
\begin{table}[b!]
	\centering
	\caption{Quantitative assessment. Average metrics on the considered subset of ImageNet-1K validation dataset for ConvNeXt Base model. Bold terms denote the best performance and underlined terms represent the second best performance.}
	\begin{tabular}{lccccc}
		\toprule
		\begin{tabular}[c]{@{}c@{}}Eval.\\ Name\end{tabular} &     FovEx   &  \begin{tabular}[c]{@{}c@{}}grad\\ CAM\end{tabular}  &   \begin{tabular}[c]{@{}c@{}}grad\\ CAM++\end{tabular}  &   \begin{tabular}[c]{@{}c@{}}Mean.\\ Pert.\end{tabular} &  \begin{tabular}[c]{@{}c@{}}random\\ CAM\end{tabular} \\ \midrule 
		\textsc{Avg. \% drop}  ($\downarrow$)    &     \textbf{59.6592}   &      80.9472    &    \underline{78.2356}   &     87.0630              &      86.6874       \\
		\textsc{Avg. \% increase} ($\uparrow$)   &     \textbf{9.7699}   &      3.4699    &    \underline{3.6999}   &     1.6999              &      0.7499       \\
		\textsc{Delete} ($\downarrow$)           &             0.2142    &      0.2098    &    \underline{0.1907}   &   \textbf{0.0739}       &      0.2638       \\
		\textsc{Insert} ($\uparrow$)             &    \textbf{0.1833}    &      0.1548    &    \underline{0.1606}   &     0.1422              &      0.1478       \\
		\textsc{EBPG}  ($\uparrow$)              &   48.1485              &\textbf{53.1640} &    \underline{49.4505}   &     43.3129              &      30.9518       \\
		\bottomrule
	\end{tabular}
	\label{tab:QuantconvNext}
\end{table}
We compare the performance of \fvx{} for three additional predictors ConvNeXt, ViT-B/16, and ViT-B/32 from torchvision. The ViT-B/16 model considered here is pre-trained on ImageNet-1K whereas the ViT-B/16 model in the main experiments is pre-trained on the ImageNet-21K dataset and fine-tuned on the ImageNet-1K dataset. We consider gradCAM \cite{selvaraju2017grad}, gradCAM++ \cite{chattopadhay2018grad}, and Mean. Pert. \cite{fong2017interpretable} for comparison in the case of the ConvNeXt Base model. For ViT-B/16 and ViT-B/32 models we consider gradCAM and gradCAM++ for comparison. We employ randomCAM as a baseline method. We do not consider RISE \cite{Petsiuk2018rise} for ConvNeXt, ViT-B/16, and ViT-B/32 due to the high time requirements to generate the attribution maps. Similarly,  due to high computational complexity, the Mean. Pert. method is not considered for ViT-B/16 and ViT-B/32. Transformer-specific methods like GAE \cite{Chefer2021ICCV} and Cls. Emb. \cite{vilas2023analyzing} methods are not utilized due to the need to change the pre-existing implementation from torchvision. 

\begin{figure}
    \centering    \includegraphics[trim={0 20cm  0 0},clip, width=0.8\linewidth]{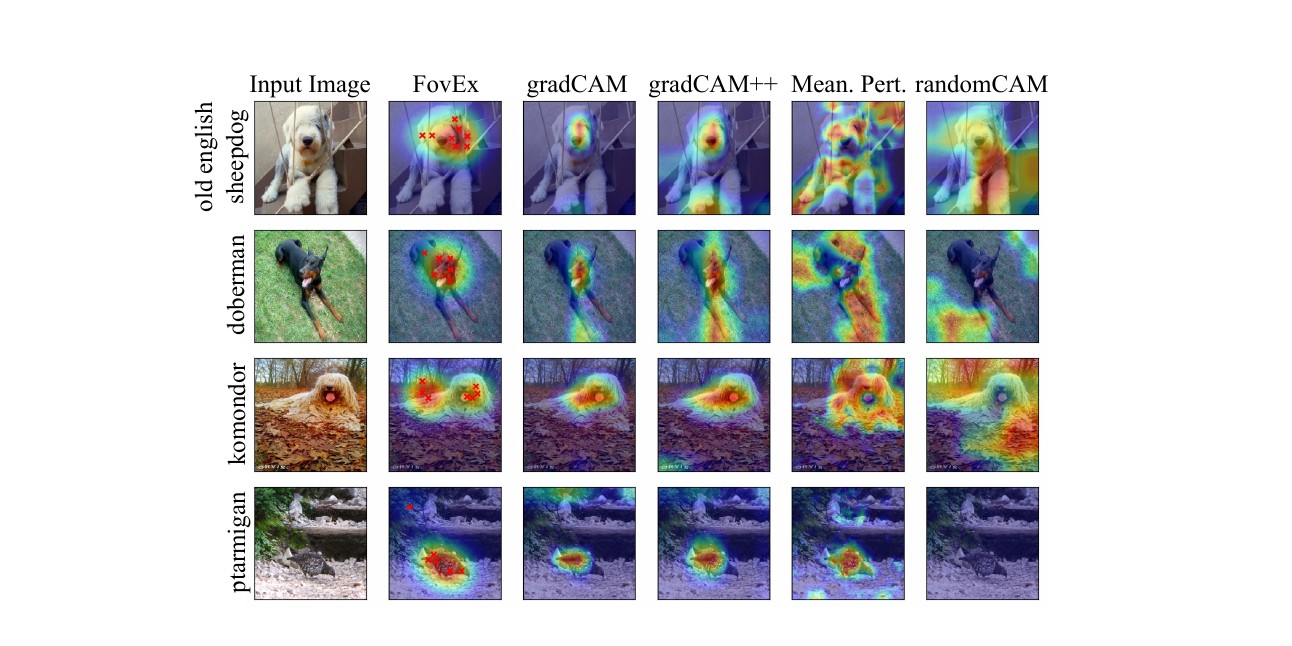}
    \caption{Qualitative assessment. Visual comparison of attribution maps generated using FovEx, gradCAM, gradCAM++, and Mean. Pert. for ConvNeXt Base model.}
    \label{fig:ConvNextQual}
\end{figure}

\begin{figure}
    \centering
    \includegraphics[width=0.5\linewidth]{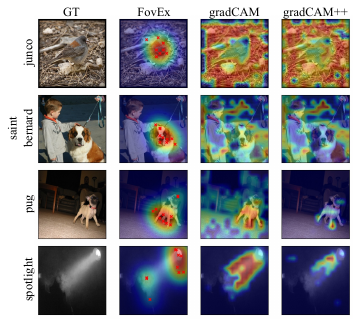}
    \caption{Qualitative assessment. Visual comparison of attribution maps generated using FovEx, gradCAM, gradCAM++, and randomCAM for ViT-B/16 model.}
    \label{fig:vitB16Qual}
\end{figure}

\begin{figure}
    \centering
    \includegraphics[width=0.5\linewidth]{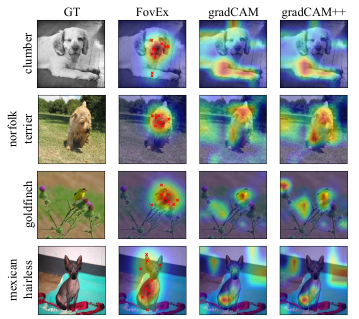}
    \caption{Qualitative assessment. Visual comparison of attribution maps generated using FovEx, gradCAM, gradCAM++, and randomCAM for ViT-B/32 model.}
    \label{fig:vitB32Qual}
\end{figure}

We report qualitative assessments in Figure \ref{fig:ConvNextQual}, Figure \ref{fig:vitB16Qual}, and Figure \ref{fig:vitB32Qual} for ConvNeXt Base, ViT-B/16 and ViT-B/32 models respectively. For all three cases, FovEx creates noise-free and consistent attribution maps. Table \ref{tab:QuantconvNext}, Table \ref{tab:vit16Quant}, and Table \ref{tab:vit32Quant} depict the qualitative evaluation results for the ConvNeXt Base model, ViT-B/16 and ViT-B/32 model respectively. FovEx maintains state-of-the-art performance across all the models by outperforming all other XAI methods in 3 out of 5 metrics in the case of ConvNeXt and in 5 out of 5 metrics in the case of ViT-B/16 and ViT-B/32 models. 

\label{performance}
\begin{table}
	\centering
	\caption{Quantitative assessment. Average metrics on the considered subset of ImageNet-1K validation dataset for ViT-B/16 model. Bold terms denote the best performance and underlined terms represent the second best performance.}
  \small
	\begin{tabular}{lccccc}
		\toprule
		\begin{tabular}[c]{@{}c@{}}Eval.\\ Name\end{tabular} &     FovEx   &  \begin{tabular}[c]{@{}c@{}}grad\\ CAM\end{tabular}  &   \begin{tabular}[c]{@{}c@{}}grad\\ CAM++\end{tabular}   &  \begin{tabular}[c]{@{}c@{}}random\\ CAM\end{tabular} \\ \midrule 
		\textsc{Avg. \% drop}  ($\downarrow$)    &     \textbf{48.2258}   &      \underline{66.9577}    &    75.6945   &     91.4881       \\
		\textsc{Avg. \% increase} ($\uparrow$)   &     \textbf{17.1490}   &      \underline{4.9299}    &    3.5299   &     0.6490       \\
		\textsc{Delete} ($\downarrow$)           &     \textbf{0.0886}   &      \underline{0.0913}    &    0.1101   &     0.1283       \\
		\textsc{Insert} ($\uparrow$)             &     \textbf{0.3647}   &      \underline{0.2956}    &    0.2964   &     0.2777       \\
		\textsc{EBPG}  ($\uparrow$)              &     \textbf{48.1485}   &      \underline{39.6095}    &    38.0989   &     36.0899       \\
		\bottomrule
	\end{tabular}
	\label{tab:vit16Quant}
\end{table}

\begin{table}[t!]
	\centering
	\caption{Quantitative assessment. Average metrics on the considered subset of ImageNet-1K validation dataset for ViT-B/32 model. Bold terms denote the best performance and underlined terms represent the second best performance.}
   \small
	\begin{tabular}{lccccc}
		\toprule
		\begin{tabular}[c]{@{}c@{}}Eval.\\ Name\end{tabular} &     FovEx   &  \begin{tabular}[c]{@{}c@{}}grad\\ CAM\end{tabular}  &   \begin{tabular}[c]{@{}c@{}}grad\\ CAM++\end{tabular}   &  \begin{tabular}[c]{@{}c@{}}random\\ CAM\end{tabular} \\ \midrule 
		\textsc{Avg. \% drop}  ($\downarrow$)    &     \textbf{44.0460}   &      \underline{64.1220}    &    70.1500              &     85.8250       \\
		\textsc{Avg. \% increase} ($\uparrow$)   &     \textbf{32.6890}   &      \underline{16.8690}    &    13.7690              &     5.3499       \\
		\textsc{Delete} ($\downarrow$)           &     \textbf{0.1034}   &      \underline{0.1261}    &    0.1241              &     0.1565       \\
		\textsc{Insert} ($\uparrow$)             &     \textbf{0.4992}   &      0.4293                &   \underline{0.4303}   &     0.3773       \\
		\textsc{EBPG}  ($\uparrow$)              &     \textbf{48.4390}   &      40.1740                &   \underline{40.9060}   &     37.6540       \\
		\bottomrule
	\end{tabular}
	\label{tab:vit32Quant}
\end{table}
\section{Default Parameter Values.}\label{secA2}
\begin{table}[h!]
	\centering
	\caption{Parameter Values. These values are used in Section \ref{imagenetExp} and Appendix \ref{secA1}.}
   \small
	\begin{tabular}{cc}
		\toprule
		Parameter Name &     Value    \\ \midrule 
		\textsc{rr} & True \\
        $\sigma_f$& 0.15 \\
        $\beta$ & 0.1 \\
        $N$ & 10 \\
        \textsc{os} & 20 \\
        $\sigma_b$ & 10 \\
        \textsc{BFS} & 41\\
        $\lambda$ & 0.1 \\
		\bottomrule
	\end{tabular}
	\label{tab:param}
\end{table}

\newpage
\section{Additional Visualizations of Explanation Maps}\label{secA3}
\begin{figure}[h!]
    \centering
    \includegraphics[width=0.6\linewidth]{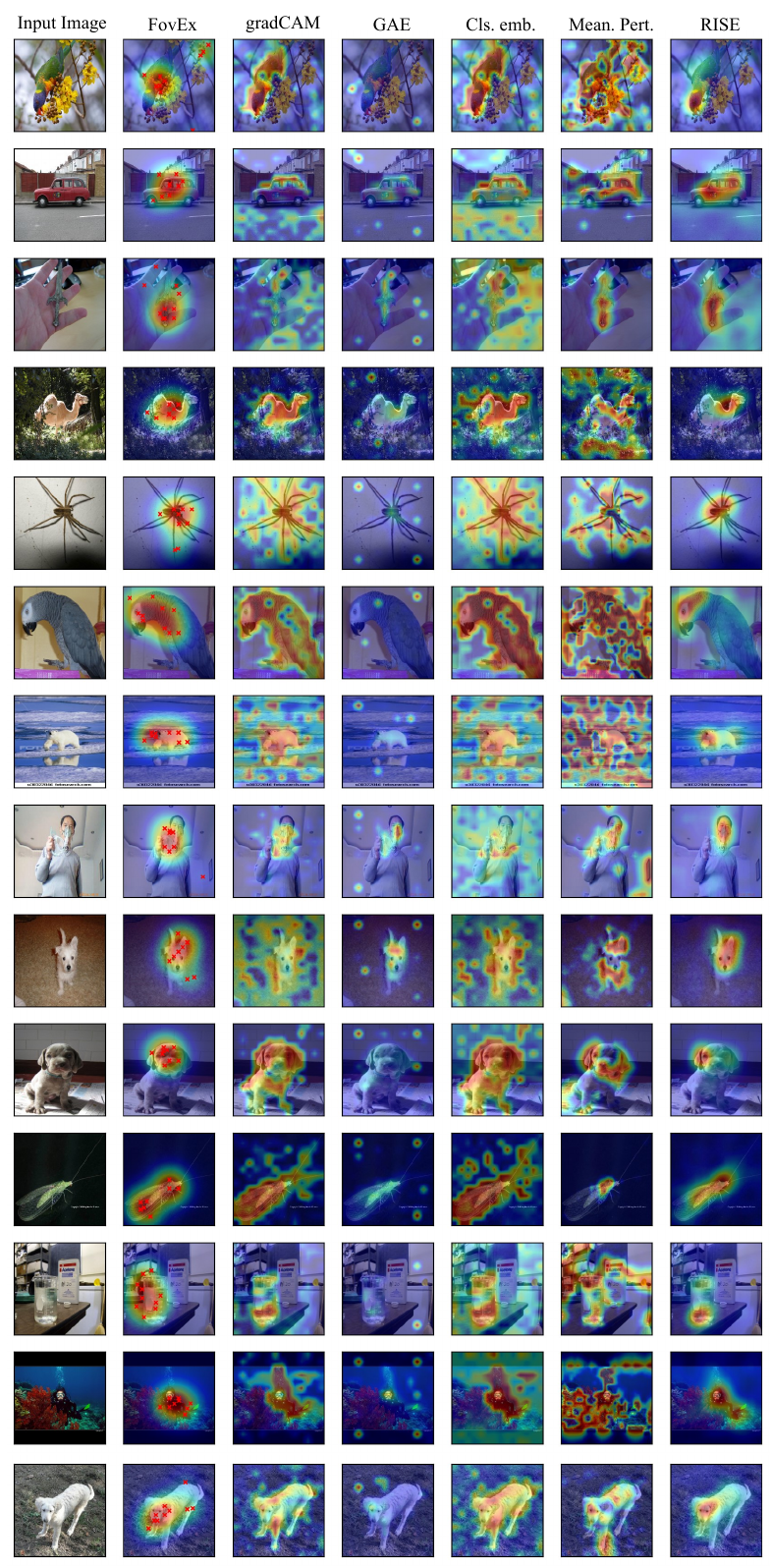}
    \caption{Qualitative assessment. Visual comparison of attribution maps generated using FovEx, gradCAM, gradCAM++, GAE, Cls. Emb., Mean. Pert., and RISE for ViT-B/16 model used in Section \ref{imagenetExp}.}
    \label{fig:vitB16All}
\end{figure}
\begin{figure}[tb!]
    \centering
    \includegraphics[width=0.7\linewidth]{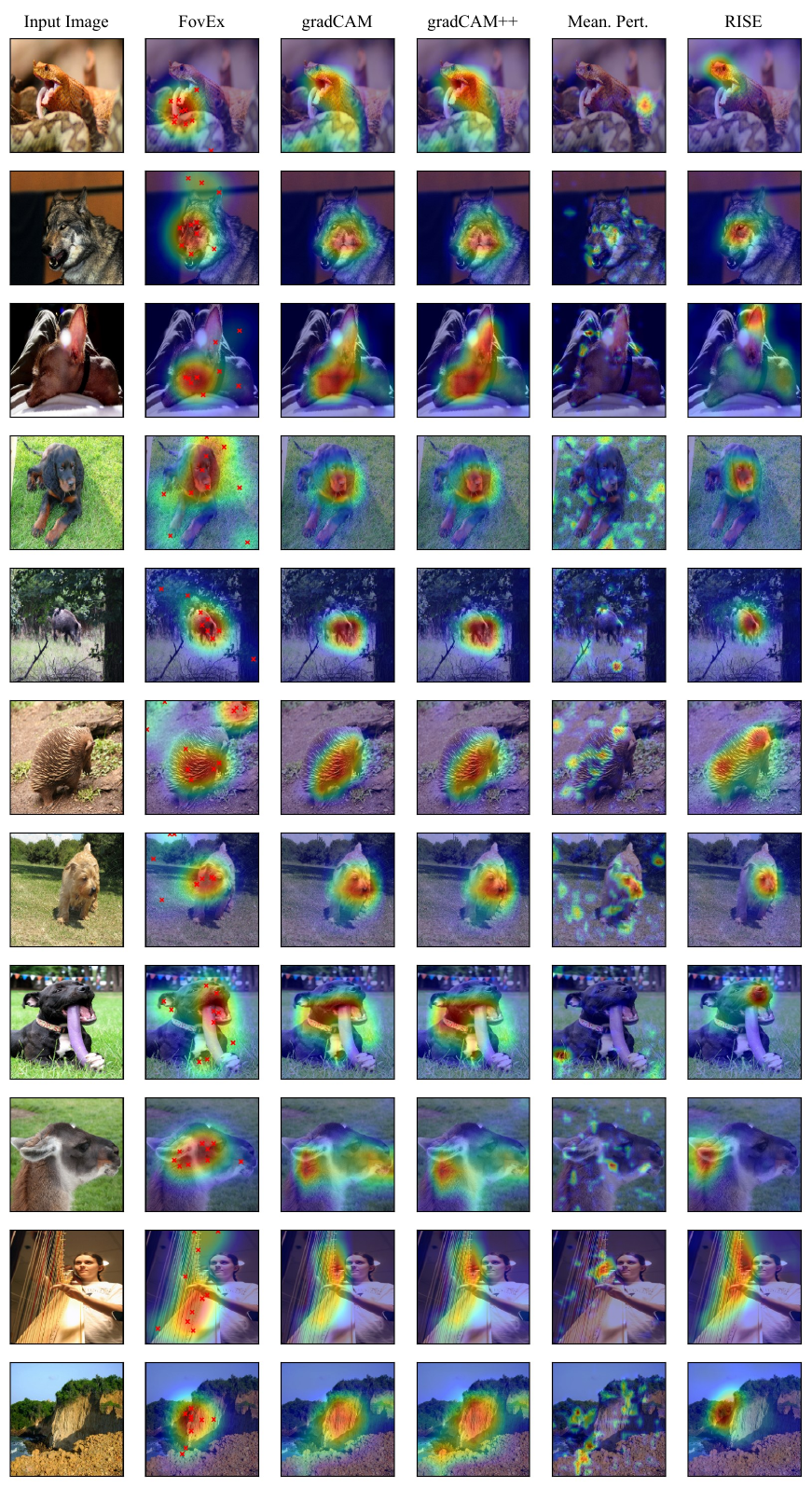}
    \caption{Qualitative assessment. Visual comparison of attribution maps generated using FovEx, gradCAM, gradCAM++, Mean. Pert., and RISE for ResNet-50 model used in Section \ref{imagenetExp}.}
    \label{fig:ResNet50All}
\end{figure}
\begin{figure}[tb!]
    \centering
    \includegraphics[width=0.5\linewidth]{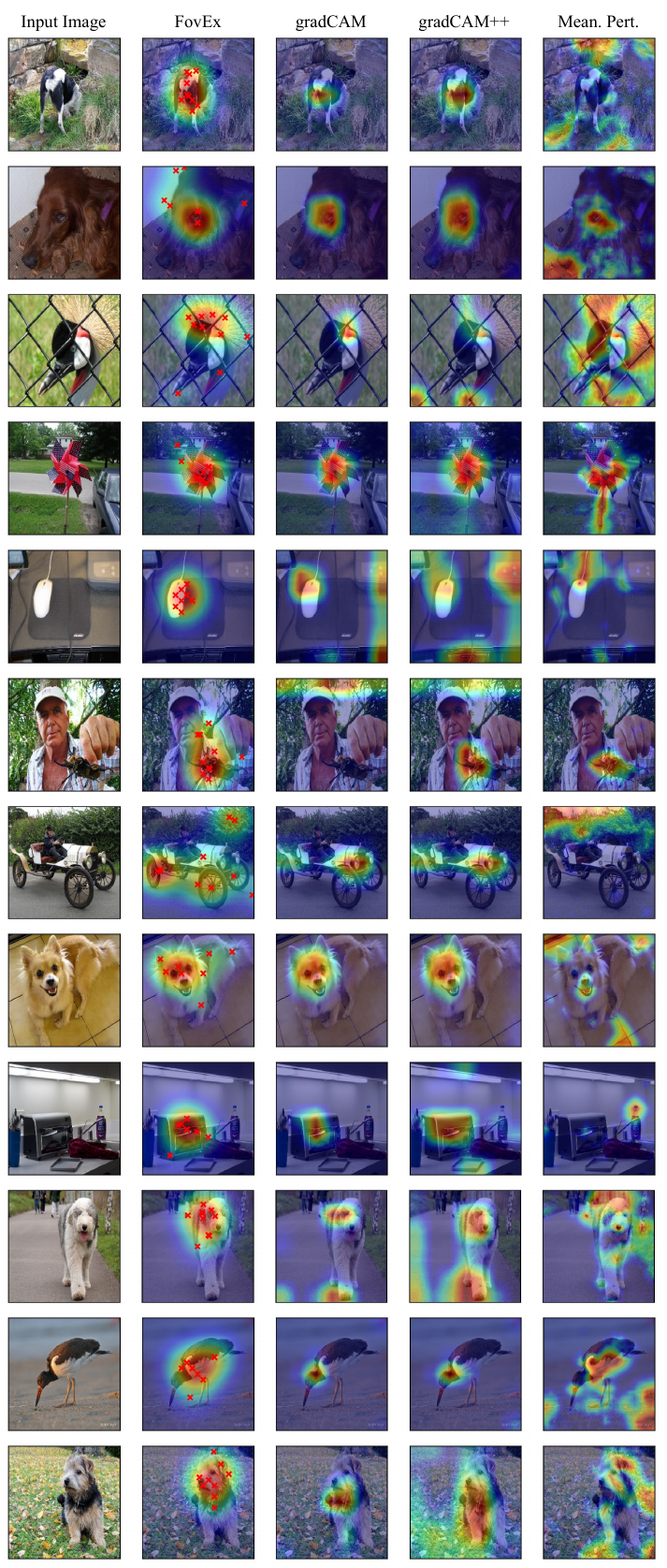}
    \caption{Qualitative assessment. Visual comparison of attribution maps generated using FovEx, gradCAM, gradCAM++, and Mean. Pert. for ConvNeXt model used in Appendix \ref{secA1}.}
    \label{fig:ResNet50All2}
\end{figure}



\end{appendices}


\end{document}